\documentclass[opre,nonblindrev]{informs4}
\RequirePackage{tgtermes}
\RequirePackage{newtxtext}
\RequirePackage{newtxmath}
\RequirePackage{bm}
\RequirePackage{endnotes}

\OneAndAHalfSpacedXII % Current default line spacing
\usepackage{algorithm}
\usepackage{algpseudocode}
\usepackage{tikz}
\usepackage{booktabs}
\usepackage{multirow}
\makeatletter
\newenvironment{breakablealgorithm}
  {% \begin{breakablealgorithm}
   \begin{center}
     \refstepcounter{algorithm}% New algorithm
     \hrule height.8pt depth0pt \kern2pt% \@fs@pre for \@fs@ruled
     \renewcommand{\caption}[2][\relax]{% Make a new \caption
       {\raggedright\textbf{\ALG@name~\thealgorithm} ##2\par}%
       \ifx\relax##1\relax % #1 is \relax
         \addcontentsline{loa}{algorithm}{\protect\numberline{\thealgorithm}##2}%
       \else % #1 is not \relax
         \addcontentsline{loa}{algorithm}{\protect\numberline{\thealgorithm}##1}%
       \fi
       \kern2pt\hrule\kern2pt
     }
  }{% \end{breakablealgorithm}
     \kern2pt\hrule\relax% \@fs@post for \@fs@ruled
   \end{center}
  }
\makeatother

\renewcommand{\enotesize}{\small} 
\usepackage{natbib}
\usepackage{xurl}
\usepackage[colorlinks=true, linkcolor=black, urlcolor=black, citecolor=black]{hyperref}
 \bibpunct[, ]{(}{)}{,}{a}{}{,}%
 \def\bibfont{\small}%

 \DeclareUrlCommand\burl{}

\EquationsNumberedThrough    % Default: (1), (2), ...
\TheoremsNumberedThrough     % Preferred (Theorem 1, Lemma 1, Theorem 2)
\ECRepeatTheorems  %  

\begin{document}
%%%%%%%%%%%%%%%%

%\setcounter{page}{1}
%\DOI{10.1287/xxxx.0000.0000}%

\RUNAUTHOR{Wang et al.}

% \RUNTITLE{Predictive Maintenance in Manufacturing}
\RUNTITLE{Non-Rival Data as  Rival Products: An Encapsulation-Forging Approach for Data Synthesis}

% \TITLE{Optimal Resource Allocation in Humanitarian Logistics: A Stochastic Programming Approach}
\TITLE{Non-Rival Data as  Rival Products: An Encapsulation-Forging Approach for Data Synthesis}
% \TITLE{Privacy-Preserving and Sample-Efficient data synthesis for Competitive Data Marketplaces}

\ARTICLEAUTHORS{%
%\AUTHOR{John Doe,\textsuperscript{a} Jane Smith,\textsuperscript{b}}
%\AFF{\textsuperscript{a}Department of Industrial Engineering, University of XYZ, \EMAIL{john.doe@xyz.edu; \textsuperscript{b}Department of Computer Science, University of ABC, \EMAIL{jane.smith@abc.edu}} 
	\AUTHOR{Kaidong Wang, Jiale Li, Shao-Bo Lin\thanks{corresponding author: sblin1983@gmail.com}, Yao Wang}
	\AFF{Center for Intelligent Decision-Making and Machine Learning, School of Management, Xi'an Jiaotong University, Xi'an, China}
} 

\ABSTRACT{
The non-rival nature of data creates a dilemma for firms: sharing data unlocks value but risks eroding competitive advantage. 
% Inter-firm data sharing is often stymied by the co-opetition dilemma, where the fear of eroding competitive advantage hinders collaboration.
Existing data synthesis methods often exacerbate this problem by creating data with symmetric utility, allowing any party to extract its value. This paper introduces the Encapsulation-Forging (EnFo) framework, a novel approach to generate rival synthetic data with asymmetric utility. EnFo operates in two stages: it first encapsulates predictive knowledge from the original data into a designated ``key'' model, and then forges a synthetic dataset by optimizing the data to intentionally overfit this key model. This process transforms non-rival data into a rival product, ensuring its value is accessible only to the intended model, thereby preventing unauthorized use and preserving the data owner's competitive edge. Our framework demonstrates remarkable sample efficiency, matching the original data's performance with a fraction of its size, while providing robust privacy protection and resistance to misuse. EnFo offers a practical solution for firms to collaborate strategically without compromising their core analytical advantage.
% Inter-firm data sharing is often stymied by the co-opetition dilemma, where the fear of proprietary knowledge leakage hinders collaboration. To address this, we propose a strategy-oriented framework for data synthesis. Our two-stage process, ``Knowledge Encapsulation'' and ``Asymmetric Utility Forging'', first encapsulates a firm's proprietary knowledge within a designated model (the ``key''), then forges a synthetic dataset with asymmetric utility, centered on that model. This resulting data artifact only reveals its full value to the key-holder, enabling secure data sharing while safeguarding competitive advantage. Our research provides a novel theory and method to transform data sharing from a strategic risk into a controllable strategic instrument.
}%

% \FUNDING{This research was supported by [grant number, funding agency].}

%Supplemental Material:
%Data Ethics & Reproducibility Note:

%\KEYWORDS{Stochastic programming, Decision support,Uncertainty, Disaster response, Optimization}

%\HISTORY{Received: Month DD, YYYY; Accepted: Month DD, YYYY; Published Online: Month DD, YYYY}

\maketitle
%%%%%%%%%%%%%%%%%%%%%%%%%%%%%%%%%%%%%%%%%%%%%%%%%%%%%%%%%%%%%%%%%%%%%%

% Text of your paper here

\section{Introduction}\label{sec:Intro}

In the contemporary digital economy, data has emerged as a strategic asset of paramount importance, fundamentally shaping how firms engage in targeted marketing \citep{marotta2022welfare}, offer personalized  product recommendations \citep{ghose2019modeling}, and examine creative idea-generation processes \citep{aggarwal2021learning}. A core economic characteristic that distinguishes data from traditional assets is its \textit{non-rivalry}  \citep{jones2020nonrivalry}, i.e., data can be used   concurrently by multiple parties without diminishing its  utility.
While non-rivalry significantly promotes inter-firm collaboration and supports the development of thriving data ecosystems, it simultaneously introduces two fundamental challenges that undermine  controllability and economic value of data during collaboration or sharing \citep{liu2025data}: \textit{data breach} and \textit{competitiveness erosion}.

% . Most physical and human resources are rival: if a person consumes a kilogram of rice or an hour of an accountant’s time, some resource with a positive opportunity cost is used up \citep{jones2020nonrivalry}. In stark contrast, . 
% enables the generation of significant network effects and returns to scale, creating value that can be broadly distributed among participants in a data ecosystem. 
% For instance, shared logistics data in supply chain management allows for collaborative predictive modeling, significantly reducing costs, minimizing delays, and lowering carbon emissions industry-wide, benefiting all participants beyond what any single firm could achieve alone; Collaborative data sharing in assortment optimization enables joint predictive analysis, leading to cost savings, reduced inventory inefficiencies, and enhanced customer satisfaction across the retail sector, surpassing the capabilities of individual companies in isolation. 
% An economic imperative for the non-rivalry is compelling, just as  
% A white paper launched by the World Economic Forum estimated that
% even in the manufacturing sector alone, the value unlocked by sharing data for process optimization could exceed \$100 billion\endnote{\burl{https://www.weforum.org/publications/share-to-gain-unlocking-data-value-in-manufacturing/}}. 
% but simultaneously    degrades  the controllability of data \citep{liu2025data}, due to which data breaches happen frequently, causing significant trouble for data owners. 

Data breach, a  widely recognized challenge stemming from the non-rivalry   of data, is closely associated with privacy risks \citep{fallah2024optimal, li2023reidentification, wang2018t} and poses serious concerns for data owners. The potential for data breaches \citep{cong2021knowledge}  has prompted the development of legal and ethical frameworks,  such as the General Data Protection Regulation (GDPR) and the California Consumer Privacy Act (CCPA),  which mandate the anonymization of personally identifiable information prior to data collaboration or sharing. However, even with anonymization, data remains vulnerable to many attacks, such as linkage attacks, attribute inference attacks, and model extraction attacks \citep{li2023reidentification}, and
 data breaches continue to occur. 
 % A typical example is the well known   Facebook-Cambridge Analytica data scandal, in which Cambridge Analytica collected the personal data of 87 million Facebook users to deploy targeted political ads designed to influence their voting decisions\endnote{\burl{https://en.wikipedia.org/wiki/Facebook-Cambridge_Analytica_data_scandal}}.
 In recent years, data synthesis has emerged as a promising solution for mitigating data breaches. A variety of approaches, such as generative adversarial networks (GANs) \citep{anand2023using}, diffusion models (DMs) \citep{sohl2015deep}, and transformer-based large language models (LLMs) \citep{zhang2023survey}, have been developed to generate artificial datasets that preserve both privacy and utilities simultaneously.

Competitiveness erosion  is a subtle yet critical  challenge, relating  to the extractable business intelligence latent within the data.
One of the key design flaws of traditional data synthesis schemes is their  {\textit{symmetric utility}, i.e., utility of data can be equally extracted   by any sufficiently advanced models.} This symmetry exposes companies to competitiveness erosion, as unauthorized access to such datasets can enable the replication of sensitive business insights, thereby threatening the company’s competitive edge \citep{liu2025data}.
This risk was vividly illustrated in the Waymo v. Uber litigation\endnote{\burl{https://www.reuters.com/article/world/waymo-accepts-245-million-and-ubers-regret-to-settle-self-driving-car-disput-idUSKBN1FT2BD/}}, in which Waymo  alleged that a former employee misappropriated over 14,000 confidential files containing proprietary LiDAR designs and test data. These materials were allegedly used by a competitor to  replicate Waymo's ``commercially sensitive insights'', thereby shortcutting years of R\&D and directly eroding Waymo's competitive advantage in the autonomous driving market.  
To   address the competitiveness erosion challenge  arising from data non-rivalry,  several solutions have been proposed that integrate synthetic data with frameworks like {\it blockchain platforms}, {\it data clean rooms}, and {\it walled gardens}, attempting to control data usage through various forms of access isolation and governance. 
Blockchain platforms, such as Ocean Protocol\endnote{\burl{https://oceanprotocol.com/}}, control data by strictly tracking and tracing its provenance, relying on contractual restrictions (often enforced via smart contracts) to govern its use cases. Similarly, data clean rooms (e.g., Snowflake\endnote{\burl{https://www.snowflake.com/en/product/features/data-clean-rooms/}}, AWS Clean Rooms\endnote{\burl{https://aws.amazon.com/cn/clean-rooms/}}) and walled gardens (e.g., Google's Ads Data Hub\endnote{\burl{https://developers.google.com/ads-data-hub}}) enforce strict usage control by physically isolating raw data and authorizing only specific, predefined functions. 
In essence, the prevailing approaches to mitigating the competitiveness erosion challenge  rely on modern forms of physical isolation.

Our approach to simultaneously addressing the data breach and competitiveness erosion challenges is  an asymmetric data synthesis strategy to yield rival data, which is inspired by four practical observations:
\begin{itemize}
\item {\bf Model-driven data requirement}: firms often pursue data sharing with the strategic goal of leveraging external data to optimize their own business models. A typical example is IBM’s acquisition of Merge Healthcare, which provides access to a large volume of medical imaging data for training its deep learning–based Watson AI platform\endnote{\burl{https://www.reuters.com/article/business/ibm-to-buy-merge-healthcare-in-1-billion-deal-idUSKCN0QB1ML/}}.
This insight motivates us to construct synthetic  data that can only be effectively utilized by  specific, intended models, thereby reducing the risk of misuse by unintended parties.
\item {\bf Asymmetric utility requirement}: existing synthesis strategies (e.g., GANs, DMs, LLMs) primarily focus on the intrinsic properties of the data  (such as privacy and statistical  distribution),  neglecting its intended use in practice, ultimately resulting in diminished utility of the synthetic data. Developing   novel synthesis paradigms to yield data of  asymmetric utility, i.e., data is exclusive to intended models but cannot be utilized by unintended models, is thus highly desired.
\item {\bf Low compatibility requirement}: the focus of existing synthesis methods on faithfully mimicking the statistical distribution of real data often leads to high compatibility, making synthetic data easily integrable with external datasets. This amplifies the risk of data breaches and further exacerbates the negative consequences of non-rivalry. Under this circumstance, synthetic data with intentionally low compatibility is required to suppress data breaches. 
\item {\bf Data minimization requirement}: data minimization,i.e., collecting and using only the amount of data strictly necessary for a specific purpose,  has become a critical consideration, driven by both regulatory mandates, such as the GDPR (Article 5(1)(c)) that explicitly  advocates for the principle of data minimization\endnote{\burl{https://eur-lex.europa.eu/eli/reg/2016/679}}, and strong economic incentives, including reduced costs for data transmission and computational model training.
\end{itemize}

In a nutshell, these  observations highlight the need for a new class of synthetic data: datasets that are compact (minimal), non-compatible (not mixable), and engineered for model-specific utility. In this way, the synthetic data ensures that even if acquired by unauthorized third parties, it cannot be used to extract accurate or commercially valuable insights. Such an approach fundamentally shifts the paradigm from  {\it physical isolation} to  {\it intrinsic technical restriction}, safeguarding the commercial value of data through the technology itself.

\begin{figure}[h]
    \FIGURE
    {\includegraphics[width=0.85\linewidth]{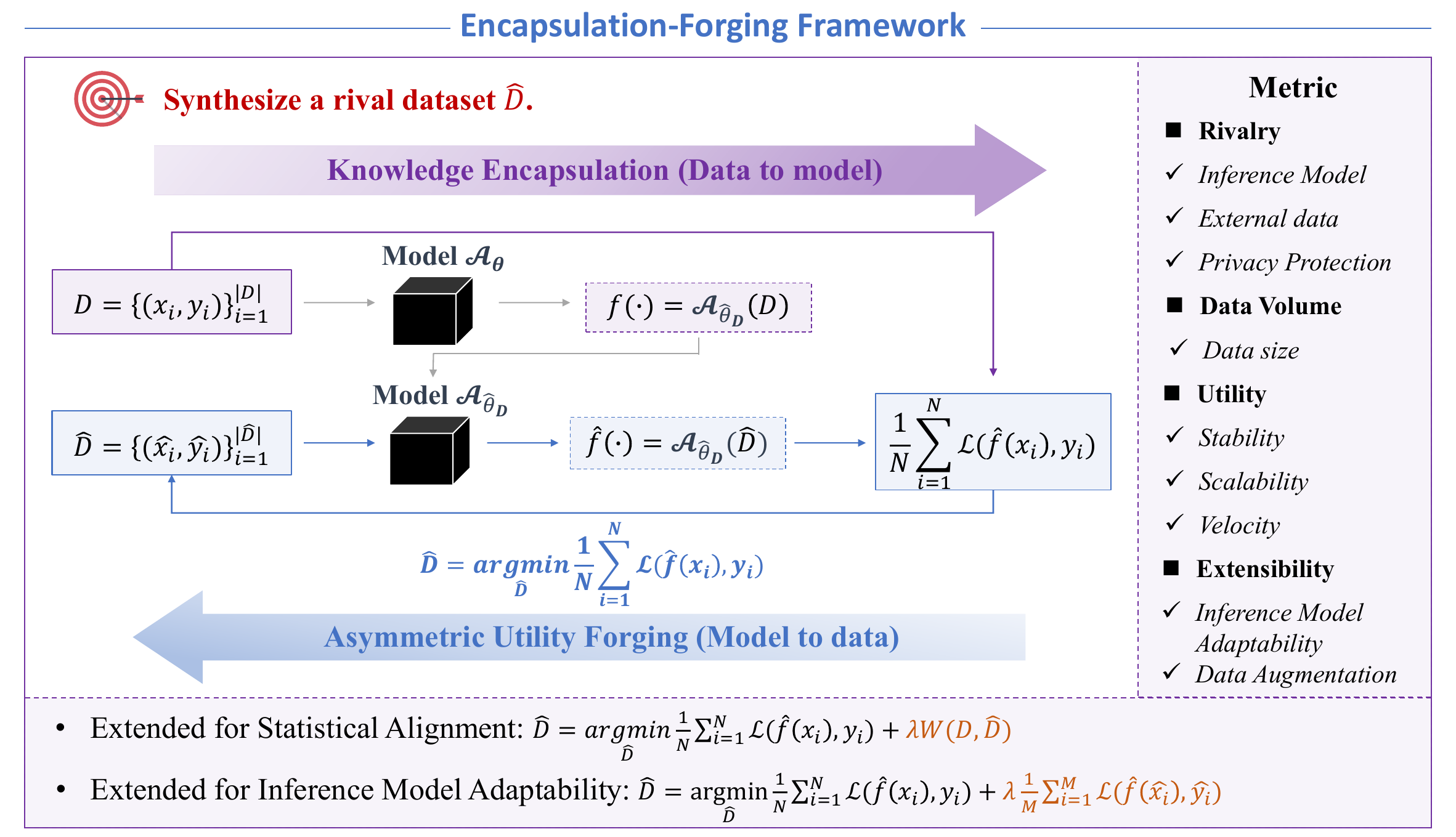}}
    {EnFo for data synthesis\label{fig:Road-map}}
    {} 
\end{figure}

This paper proposes a novel {\it encapsulation-forging} (EnFo) framework for synthesizing data with asymmetric utility. The framework consists of two key stages:  {\it knowledge encapsulation} that distills predictive patterns from the original data  into a intended  model, and {\it asymmetric utility forging} which crafts the synthetic data by optimizing the data   to minimize  empirical risk over the encapsulated model.
% To synthesize asymmetric data, firms need  to absorb the model information into the synthesization process. Thus, this paper propose an EnFo framework to synthetic data with asymetric utility, via encapsulating the knowledge   of   data into a specific model and then forging the synthetic data through implementing empirical risk minimization (ERM) on the encapsulated model.
%This paper aims to propose a brand-new paradigm for   data generation to enable synthetic data to act as   rival products via encapsulating the knowledge   of   data into a specific model and then forging the synthetic data through implementing empirical risk minimization (ERM) on the encapsulated model. 
% Different from classical machine learning, which applies empirical risk minimization (ERM) to train a model that avoids over-fitting (i.e.,  the model fits the training data too well while failing to generalize to new observations), 
EnFo  aims to generate data that aligns exclusively with a  intended model, requiring the synthetic data to over-fit the model, which is different from the classical data fitting and machine learning devoting to avoid over-fitting. 
On this basis, EnFo succeeds  in synthesizing rival  data with asymmetric utility, wherein the data is specifically tailored to the intended model and provides little to no value when used with others. Notably, to ensure overfitting while simultaneously preserving utility-dependent privacy, the size of data synthesized by EnFo should be significantly smaller   than that of the original dataset, which directly aligns with the data minimization requirement.
To broaden the applicability of EnFo across different models and datasets, two regularization schemes on statistical alignment and model adaptability are   proposed. 
The core idea and workflow of EnFo are illustrated in Figure \ref{fig:Road-map}. 

Our study makes several key contributions to the methodology and practice of data synthesis.
On the methodological front, we introduce a novel, strategy-oriented approach that replaces the statistical mimicry of traditional models (e.g., GANs, DMs, and LLMs) with a model-driven EnFo framework that directly optimizes synthetic data for an intended model. By encapsulating the information of the original data   into the intended model, we forge synthetic data that exhibits superior  performance exclusively on this model through a benign over-fitting alignment. This transforms non-rival  data into rival synthetic data tailored for specific scenarios. Moreover, EnFo incorporates a flexible regularization mechanism that enables precise control over the exclusivity and fidelity of the resulting  `` data lock''.

% This approach allows us to pioneer the concept of ``controllable utility'', where data providers can precisely dictate which models benefit from the data to prevent unauthorized exploitation. Furthermore, we develop a flexible regularization mechanism for granular control over the ``knowledge lock's'' specificity, privacy, and data fidelity. Our method also achieves remarkable sample efficiency, generating compact datasets that are significantly smaller than the original yet achieve comparable or superior performance, addressing practical storage and transmission constraints.

Our extensive empirical validation, conducted across five diverse datasets and eleven metrics, confirms the superior performance of the EnFo framework. We demonstrate its effectiveness in promoting data rivalry: EnFo degrades the performance of unauthorized models by over 80\% (Friedman dataset), and exhibits strong resistance to external data augmentation, with prediction errors increasing by up to 32\% when merged with external data. 
Beyond its core capabilities, our framework demonstrates remarkable sample efficiency, matching the predictive accuracy of the full original dataset using as little as 1\% of its volume (CLV dataset). Finally, EnFo delivers robust privacy protection, achieving Monte Carlo Attack Accuracy scores consistently below 0.57 across all experiments, substantially outperforming both traditional methods and modern generative models.

% We achieve substantial predictive improvements (35\% RMSE reduction on Instacart dataset) while using only 5\% of the original data size. Our synthetic data exhibits strong competitiveness through statistical divergence metrics that are 6-7 times higher than baselines, inference model specificity that degrades unauthorized model performance by up to 32\%, and resistance to external augmentation that prevents performance improvements from data injection. The framework also provides excellent privacy protection with Monte Carlo Attack Accuracy values consistently below 0.53, outperforming both traditional perturbation methods and modern generative models. These results validate our theoretical framework and demonstrate its practical effectiveness across diverse application domains.

From a practical and managerial perspective, EnFo offers a compelling solution to the data-sharing dilemma by transforming data into a controllable strategic asset. Our framework generates synthetic data with rivalry, ensuring its value is exclusively accessible to intended partners for specific, pre-defined purposes. This empowers firms to confidently monetize their data assets and engage in strategic collaborations without compromising their core analytical advantage. Operationally, EnFo's remarkable sample efficiency, matching full dataset performance with a fraction of the data size, provides a distinct competitive edge by reducing data management costs and accelerating time-to-insight.

The rest of the paper is organized as follows. Section \ref{sec: Literature} introduces the problem setting and reviews the related literature. Sections \ref{sec:Model} and \ref{sec: model extensions} present our proposed EnFo framework for rival data synthesis and its extensions. Section \ref{sec:Methodology} outlines a comprehensive evaluation methodology.  Section \ref{sec:Experiments} presents extensive empirical analyses based on both Monte Carlo and real-world datasets. Finally, Section \ref{sec:Conclusion} concludes the study and discusses potential directions for future research.

\section{Problem Setting and Related Work}\label{sec: Literature}

\subsection{Problem setting}\label{Problem Setting}
In modern marketing, customer data has become one of the most critical strategic assets for firms. 
Consider a firm possessing a dataset $\mathcal D= \{(x_i, y_i)\}_{i=1}^N$, where each record captures information about an individual customer. 
For each customer $i$, $x_i$ is a vector of observable characteristics (e.g., demographics, browsing history, past purchases), and $y_i$ represents a key performance outcome of interest, such as future spending or customer lifetime value (CLV). 
The underlying relationship between the input and outcome  can be captured by \citep{gyorfi2006distribution}:
\begin{equation}\label{model:regression}
    y_i = f^\star(x_i) + \varepsilon_i,
\end{equation}
where $f^\star$ denotes the unknown ground-truth  that maps customer characteristics to the outcome, and $\varepsilon_i$ captures the random noise introduced during data collection.
The fundamental value of the dataset $\mathcal D$ lies in its capability to produce an accurate estimate of  $f^\star$.  
A precise predictive model $f$ enables the firm to better understand consumer behavior, facilitating effective segmentation, targeting, and personalization, thereby fostering a sustainable competitive advantage. Mathematically, we evaluate the utility (value) of $\mathcal{D}$ under a given algorithm $\mathscr{A}_\theta: \mathcal{D} \to f$ with hyperparameters $\theta$ by
\begin{align}
    \mathcal{U}(\mathscr{A}_\theta(\mathcal{D}))=\|f_{\mathcal{D},\theta}^{\mathscr{A}}-f^\star\|,
\end{align}
where $f_{\mathcal{D},\theta}^{\mathscr{A}}:=\mathscr{A}_\theta(\mathcal{D})$ denotes the predictive model trained on $\mathcal{D}$ using algorithm $\mathscr{A}_\theta$, and $\|\cdot\|$ represents a norm measuring the distance between the learned model and the ground-truth.

The utility (value) of data is threatened by its inherent nature as a non-rival product. 
% Unlike physical product, data can be copied and used by multiple parties simultaneously without depletion. 
When a firm shares its synthetic dataset $\mathcal{\widehat{D}}$, for example  with a research partner, a data intermediary, or within a co-opetition arrangement, it inevitably loses control over its subsequent use. 
Once data $\mathcal{\widehat{D}}$ is disclosed by the recipient to a third party, the latter can exploit its utility (value) in two primary ways, thereby diluting the original owner's business value:
\begin{itemize}
    \item \textbf{Data Augmentation:} A third party can merge $\mathcal{\widehat{D}}$ with its own proprietary data $\mathcal D'$ for data augmentation \citep{guo2025selling}, which unlocks predictive power  that was inaccessible to the original owner alone via informational synergies. Competitiveness erosion consequently happens in terms of $\mathcal{U}(\mathscr{A}_\theta(\mathcal{\widehat{D}}\cup \mathcal D'))<\mathcal{U}(\mathscr{A}_\theta(\mathcal{\widehat{D}}))$.
    \item \textbf{Model Augmentation:} A third party is free to apply any inference model $\mathscr{B}_\phi$ with parameter $\phi$ to the data, including more sophisticated or powerful models (e.g., deep neural networks) than those used by the data owner, resulting in higher predictive utility such that $\mathcal{U}(\mathscr{B}_\phi(\mathcal{\widehat{D}})) < \mathcal{U}(\mathscr{A}_\theta(\mathcal{\widehat{D}}))$.
\end{itemize}

This uncontrolled proliferation of insights allows third parties (including competitors) to extract  the same or even superior value from the data,  severely diluting  the uniqueness of the data  and erode the   competitive edge it was originally intended to provide to its owner.
  Asymmetric utility  in data synthesis   is thus required 
  to  satisfy the requirements of low compatibility (LC) and model specificity (MS) to safeguard against value dilution caused by unauthorized use,
 which can be formulated as the following optimization problem:
\begin{align}
    &\argmin\limits_{\mathcal{\widehat{D}}\in \mathcal{X}\times\mathcal{Y}}~\mathcal{U}(\mathscr{A}_\theta(\mathcal{\widehat{D}})) \label{target} \\
    s.t.~  
    \mathcal{U}(\mathscr{A}_\theta&(\mathcal{\widehat{D}}\cup \mathcal D')) -\mathcal{U}(\mathscr{A}_\theta(\mathcal{\widehat{D}})) > 0,~\forall \mathcal D'\in \mathcal{X}\times\mathcal{Y} \tag{LC}\label{Constraint:Low Compatibility} \\
    \mathcal{U}(\mathscr{B}_\phi&(\mathcal{\widehat{D}})) -\mathcal{U}(\mathscr{A}_\theta(\mathcal{\widehat{D}})) > 0 , ~\forall  \mathscr{B}_\phi\neq\mathscr{A}_\theta  \tag{MS}\label{Constraint:Model Specificity}.
\end{align}
The (\ref{Constraint:Low Compatibility}) constraint specifies that the utility (value) of the shared synthetic data  $\mathcal{\widehat{D}}$ significantly degrades when merged with unauthorized external data.
The (\ref{Constraint:Model Specificity}) constraint requires that the synthetic data  $\mathcal{\widehat{D}}$ maintains high utility only when used with a pre-specified inference model $\mathscr{A}_\theta$ and using other models  leads to a substantial drop in predictive accuracy.
Specifically, $\mathcal{\widehat{D}}$ is said to be 
  \textit{rivalry} if it simultaneously satisfies both the model specificity constraint (\ref{Constraint:Model Specificity}) and the low compatibility constraint (\ref{Constraint:Low Compatibility}).
By engineering the data to retain value exclusively under prescribed conditions, firms can confidently share their data assets while safeguarding their core value against unintended or unauthorized exploitation.

 {As (\ref{target}) is applied to any data set, the (\ref{Constraint:Low Compatibility}) constraint naturally holds and is thus removable in the optimization problem.} Moreover, constraint (\ref{Constraint:Model Specificity}) implies that the synthesized data are exclusive to the model $\mathcal A_\theta$ and cannot be well generalized to other models, indicating that the data should be deliberately over-fitted to $\mathcal A_\theta$. In contrast to the classical data-to-model machine learning paradigm, where over-fitting should be avoided to enhance the generalization performance of a model, over-fitting becomes essential in the model-to-data synthesis paradigm to ensure the rivalry of the synthetic data.
To this end, the constraint (\ref{Constraint:Model Specificity}) can be reformulated as a restriction on the size of the synthetic data  $\mathcal{\widehat{D}}$, ensuring the occurrence of the desired over-fitting. We therefore relax the original optimization problem as follows:
\begin{align}\label{target-111}
    \argmin\limits_{\mathcal{\widehat{D}}\in \mathcal{X}\times\mathcal{Y},|\widehat{D}|\leq M}~\mathcal{U}(\mathscr{A}_\theta(\mathcal{\widehat{D}})),
\end{align}
where $M\in\mathbb N$ is a pre-specified parameter much smaller than $|\mathcal{D}|$ to guarantee that $\mathcal A_\theta$ over-fits $\mathcal{\widehat{D}}$. This relaxation enables the synthesis of rival data with asymmetric utility and is the main purpose of our study.

\subsection{Related work}
In this part, we introduce several related work and conduct some comparisons.
\subsubsection{Non-rivalry of Data Product}
Our work builds on the literature that conceptualizes data as a non-rival economic resource and examines the strategic tensions arising from this characteristic.
A number of studies have emphasized that data can be simultaneously used by multiple firms without depletion, highlighting the potential for significant social benefits through data sharing \citep{jones2020nonrivalry, farboodi2023data}. 
These studies, however, also identify a central dilemma in generating rival data.
For instance, in their macroeconomic model, Jones and Tonetti (2020) demonstrated that while broad data access is socially optimal, firms are incentivized to hoard proprietary data to mitigate the risk of ``creative destruction'' from rivals, which results in an inefficiently narrow use of such non-rival resource.
Some studies have examined the potential effects of mandatory data-sharing policies. 
For example, a study by \citep{kramer2025regulating} showed that while data sharing can spur innovation by fostering competition, it may also dampen the innovation incentives of incumbent firms due to a ``loss of exclusivity". \cite{liu2025data} further elaborated on this disincentive, noting that it becomes more pronounced when data act as ``strategic substitutes", where information known by others diminishes its own value.
% The work also identifies a central dilemma: while sharing is socially optimal, firms are incentivized to hoard data to prevent ``creative destruction'', leading to market inefficiencies \citep{jones2020nonrivalry}. Recent studies on mandatory data-sharing policies show that while such regulations can foster competition, they may also dampen innovation incentives due to a ``loss of exclusivity'' \citep{kramer2025regulating}, a disincentive amplified by the “strategic substitutability” of data, where its value diminishes if competitors have it too \citep{liu2025data}.

% \cite{jones2020nonrivalry} develop a macroeconomic model to analyze these implications, contrasting different property rights regimes. They demonstrate that when firms own data, they hoard it to mitigate the risk of creative destruction, leading to an inefficiently narrow use of this nonrival resource. As a potential solution, the paper suggests that assigning data property rights to consumers can achieve a nearly optimal allocation. This framework empowers consumers to balance their privacy concerns against the economic gains from selling data broadly, thereby unlocking the significant social value inherent in its nonrival nature.

Compared to this rich body of literature, our contribution lies in offering a novel approach to addressing the strategic dilemma of data sharing. In contrast to prior studies that frame the issue as a binary choice between hoarding and sharing existing data, our approach provides a useful counterpoint by fundamentally altering the properties of the shared data itself. We introduce a data synthesis technique that generates utility-controllable datasets, where the synthetic data are engineered to be highly effective only when paired with specific proprietary algorithms, while exhibiting low utility for unintended users. This approach enables firms to engage in value-enhancing data sharing while strictly controlling the scope and recipients of data usage, directly addressing core issues such as creative destruction and value dilution that underlie data hoarding.

\subsubsection{Co-opetition in Data Sharing}
The strategic dilemma inherent in data sharing is a central concern in the literature on co-opetition, which examines how firms simultaneously cooperate and compete in data-rich environments. 
\citep{chen2025r} highlights a fundamental tension between corporate strategies of selective sharing and regulatory goals of comprehensive data access.
At its core, this tension arises from what \cite{liu2025data} characterized as the ``strategic substitutability'' of data, where sharing with a competitor reduces the relative value of a firm's proprietary information assets. \cite{liu2025data} further formalized this disincentive at the microeconomic level as ``data dilution'', showing that a data seller inherently competes with its future self: the inability to guarantee  exclusivity (i.e., ensuring the data won’t be sold to others later) undermines the data's value for early buyers. This dynamic creates what \cite{kramer2025regulating} termed an ``exclusivity paradox'': mechanisms designed to increase collaboration by reducing exclusivity can inadvertently destroy the very incentive to create valuable data in the first place.
While mandated sharing may be beneficial for overall welfare, this ``loss of exclusivity'' has been shown to  reduce innovation incentives for incumbent firms. From the firm's perspective, the incentive to share data voluntarily is highly conditional. \cite{chen2025r} demonstrated that firms are more likely to share when data is complementary or when market uncertainty is high, as the benefit of growing the entire market outweighs the loss of competitive advantage.

Our research addresses this paradox by introducing what we term ``controlled asymmetric sharing'', which enables firms to collaborate while maintaining strategic control over how their shared data is used. This approach aligns with calls for sharing mechanisms that preserve competitive incentives while capturing collaborative value \citep{johnson2024online}, thereby offering a resolution to a core co-opetition dilemma in data markets.

% Our research also contributes to the growing literature on co-opetition in digital ecosystems, which examines how firms simultaneously cooperate and compete in data-rich environments. Recent work by \cite{chen2025r} highlights the fundamental tension between corporate and governmental data-sharing strategies, where firms prefer selective sharing to maintain competitive advantages while regulators advocate for comprehensive data sharing to foster innovation. This tension reflects what \cite{liu2025data} characterize as the "strategic substitutability" of data—where sharing reduces the relative value of a firm's information assets, creating disincentives for collaboration despite potential social benefits. The challenge is further complicated by what \cite{kramer2025regulating} identify as the "exclusivity paradox" in data sharing: while mandatory sharing policies can enhance competition and innovation, they may simultaneously reduce firms' incentives to invest in data collection and curation due to diminished appropriability. 

% Our framework addresses this paradox by enabling what we term "controlled asymmetric sharing"—allowing firms to collaborate while maintaining strategic control over how shared data can be utilized. This approach aligns with recent insights from \cite{johnson2024online} on the importance of designing sharing mechanisms that preserve competitive incentives while capturing collaborative value, thereby resolving the fundamental co-opetition dilemma in data markets.

\subsubsection{Data Synthesis}
The data synthesis is a central theme in the privacy-preserving literature. The field has evolved from traditional statistical perturbation and suppression methods \citep{li2011protecting, fallah2024optimal, li2023reidentification, wang2018t} to modern generative models like GAN \citep{goodfellow2014generative} and DM \citep{sohl2015deep}, which excel at creating high-fidelity, privacy-preserving datasets \citep{anand2023using, zhu2024synthetic, villaizan2025diffusion}. 
Among existing approaches, a particularly notable one is the GAN-based method proposed by \citep{anand2023using}. In this approach, only the trained generative model is transferred to the buyer—without any real or synthetic data leaving the firm—thereby providing strong privacy protection while substantially reducing data transfer costs. Moreover, the mini-batch stochastic gradient descent (SGD) based training procedure of GANs provides notable advantages in scalability and velocity, which are particularly beneficial for applications involving large-scale or continuously arriving data. Nonetheless, this approach does little to reduce the workload on the user side, since achieving comparable predictive performance usually requires generating synthetic data at full original scale. More critically, it offers no safeguard against misuse or unintended commercial exploitation. Once the generator is transferred, it can be  used repeatedly to produce unlimited synthetic data beyond its intended purpose, without any oversight or control from the data owner.

% \textcolor{blue}{However, existing methods primarily focus on two objectives: maximizing statistical fidelity and ensuring privacy. While effective at ensuring statistical fidelity and privacy, existing methods, by design, do not include specific components to preserve or modulate data utility.}
% These methods did not take utility into account during the design process, that is, no specific components were designed to protect or control utility.

Our research diverges fundamentally from this existing paradigm by introducing the concept of asymmetric data synthesis.  Unlike the above conventional methods that seek to learn distribution with symmetric universal utility, our approach is designed to synthetic data with a controlled, asymmetric specific utility. This represents a shift from a privacy-centric to an utility-centric paradigm, where the objective is not just protecting privacy but enabling model-specific utility which avoid competitiveness erosion.

\section{Encapsulation-Forging Framework}\label{sec:Model}

In this section, we present the EnFo framework for data synthesis. 
The core methodology involves two sequential stages. The first stage, {\it knowledge encapsulation}, utilizes the original data to train a specific inference model, thereby embedding the underlying predictive knowledge within it. The second stage, {\it asymmetric utility forging}, involves crafting a synthetic dataset that is intentionally engineered to overfit this pre-trained inference model. This targeted overfitting  serves as the key mechanism that induces data rivalry, making the synthetic data  highly effective for its intended user but less so for others. 
% This section set in a scenario where a data recipient has a fixed inference model but no data. The shared dataset is therefore used as a standalone asset, with its served  inference model has pre-specified.
% The core idea is through use original  data to fit the specific inference model  to encapsulate the underlying knowledge into the inference model, and overfitting the synthetic dataset to the inference model to finish the asymmetric utility forge.

\subsection{Knowledge encapsulation: fitting inference model to original  data}

In marketing practice, data sharing typically serves a specific downstream purpose: enabling partners to build predictive models for customer behavior analysis. For instance, when an e-commerce platform shares customer transaction data with a logistics partner, the primary objective is to enable accurate demand forecasting using predetermined regression models. Similarly, when a retailer collaborates with a marketing agency, the shared customer behavioral data is intended to be used with specific machine learning algorithms, such as logistic regression for churn prediction or neural networks for customer lifetime value estimation, that both parties have agreed upon in advance.

Formally, we denote the original dataset as $\mathcal{D} = \{(\mathbf{x}_i, y_i)\}_{i=1}^N$, where $\mathbf{x}_i \in \mathbb{R}^d$ represents the feature vector of the $i$-th observation (e.g., customer demographics, purchase history) and $y_i$ is the corresponding outcome variable (e.g., future spending, churn probability). 
We formalize this by assuming the specific inference model is trained using a specific algorithm 
$$
\mathscr{A}_{\theta}: \mathcal{D} \mapsto f(\cdot),
$$
which is configured by a set of hyperparameters $\theta$ and maps a given training dataset to a corresponding predictive model, $f(\cdot)$, such that
$
f(\cdot) = \mathscr{A}_{\theta}(\mathcal{D}).
$
The algorithm $\mathscr{A}_{\theta}$ can represent any standard technique in marketing analytics, from linear models (e.g., linear or ridge regression) and kernel methods to complex deep neural networks.

The first step of our framework is   knowledge encapsulation, where we distill the predictive relationships from  $\mathcal{D}$ into a definitive key model. This is achieved by  identifying the optimal hyperparameters, $\hat{\theta}_\mathcal{D}$, by training and validating on  $\mathcal{D}$ (e.g., via cross-validation). This optimization procedure systematically navigates the bias-variance trade-off to find an ideal balance: a model complex enough to capture the underlying predictive signals without overfitting to the dataset's idiosyncratic noise. The selected hyperparameters, $\hat{\theta}_\mathcal{D}$, are those  minimizing the expected generalization error and thus maximize predictive utility on unseen data. We refer to  $\mathscr{A}_{\hat{\theta}_\mathcal{D}}$ as  the key algorithm,  as it   encapsulates the essential predictive relationships from the original data.  The implementation of knowledge encapsulation varies depending on the model type:
% Based on this, we enter into Knowledge Encapsulation stage. 
% We first train $\mathscr{A}_{\theta}$ on the original  dataset $\mathcal{D}$ to obtain the optimal parameter estimate $\hat{\theta}_\mathcal{D}$. We refer to $\mathscr{A}_{\hat{\theta}_\mathcal{D}}$ as the "key" algorithm, which encapsulates the essential knowledge from the original  data. This process can be understood as distilling the predictive relationships inherent in $\mathcal{D}$ into a specific algorithmic structure. The implementation of knowledge encapsulation varies depending on the model type:
\begin{itemize}
    \item \textbf{Algorithms without hyperparameter ($\theta=\varnothing$)}: For models like standard linear regression,  $
    f(\mathbf{x}) = \mathbf{w}^\top\mathbf{x}
    $, the objective is to find the model parameters $\mathbf{w}$ that minimize the loss function: $\min_{\mathbf{w}}\frac{1}{N}\sum_{i=1}^N(f(\mathbf{x}_i)-y_i)^2$.
    The learning algorithm is fixed, applying it to   $\mathcal{D}$ yields the well-known closed-form solution for the optimal parameters $\mathbf{w}^*=(\mathbf{X}^\top\mathbf{X})^{-1}\mathbf{X}^\top\mathbf{y}$, where $\mathbf{X} \in \mathbb{R}^{N \times d}$ is the design matrix with rows $\mathbf{x}_i^\top$, and $\mathbf{y}=[y_1,y_2,\cdots,y_N]^\top \in \mathbb{R}^N$. Since no hyperparameter exists, the encapsulation is direct: the key algorithm is simply $\mathscr{A}: \mathcal{D} \mapsto f(\mathbf{x})=((\mathbf{X}^\top\mathbf{X})^{-1}\mathbf{X}^\top\mathbf{y})^\top\mathbf{x}$ itself. %clearly, there is no hyperparameter.$\mathscr{A}$ is 
    \item \textbf{Algorithms with hyperparameter ($\theta\neq\varnothing$)}: For advanced models like neural networks or kernel methods,  encapsulation is a two-step process. At first, we determine the optimal hyperparameters $\hat{\theta}_\mathcal{D}$ (such as the number of iterations and neurons) from the original  data. Then, we use these hyperparameters to define the key model $\mathscr{A}_{\hat{\theta}_\mathcal{D}}$.  For example, consider the $\nu$-method \citep{gerfo2008spectral}, an accelerated version of kernel-based gradient descents,  the inference model is given by 
    $
    	f(\mathbf{x}) = \sum_{i=1}^{N}\boldsymbol{\alpha}^{\nu,T}_iK(\mathbf{x},\mathbf{x}_i) 
    $
    % \begin{equation}\label{nu method}
    % 	\widehat{f}(\mathbf{x}) = \sum_{j=1}^{M}\boldsymbol{\alpha}_jK(\mathbf{x},\widehat{\mathbf{x}}_j),
    % \end{equation}
    with  a Mercer kernel $ K(\cdot, \cdot) $, where $\boldsymbol{\alpha}$ denotes the parameter vector. The parameters $\boldsymbol{\alpha}$ are found by an iterative process governed by a key hyperparameter, $\nu$, which controls acceleration, and the total number of iterations, $T$. The iterative update rule is (with initialization $ \boldsymbol{\alpha}^{\nu,0} = \boldsymbol{\alpha}^{\nu,-1}  = \mathbf{0} $):
    \begin{equation}\label{nu method alpha}
    	\boldsymbol{\alpha}^{\nu,t} :=g_t(\boldsymbol{\alpha}^{\nu,t-1}, \boldsymbol{\alpha}^{\nu,t-2})= \boldsymbol{\alpha}^{\nu,t-1} + u^{\nu,t}(\boldsymbol{\alpha}^{\nu,t-1} - \boldsymbol{\alpha}^{\nu,t-2}) + \frac{\omega^{\nu,t}}{|D|}(\mathbf{y} - \boldsymbol{K} \boldsymbol{\alpha}^{\nu,t-1}),~ t=1,2,\ldots,T
    \end{equation}
    where $ \boldsymbol{K}= \{K(\boldsymbol{x}_i,\boldsymbol{x}_j)\}_{i,j=1}^N$ is the kernel matrix,  $u^{\nu,t}$ and $\omega^{\nu,t}$ denote respectively the momentum and step-size coefficients at iteration $t$, analytically defined by
    \begin{equation}\label{nu method alpha2}
	\begin{aligned}
		u^{\nu,t} = \frac{(t-1)(2t - 3)(2t + 2\nu - 1)}{(t + 2\nu - 1)(2t + 4\nu - 1)(2t + 2\nu - 3)}, ~~~
		\omega^{\nu,t} = 4 \cdot \frac{(2t + 2\nu - 1)(t + \nu - 1)}{(t + 2\nu - 1)(2t + 4\nu - 1)}.
	\end{aligned}
    \end{equation}
    Here, knowledge encapsulation involves finding the optimal hyperparameters $\hat{\theta}_\mathcal{D}=\{\nu^*,T^*\}$ by training and validating on the original  data  $\mathcal{D}$ with the target $\min_{\mathbf{w}}\frac{1}{N}\sum_{i=1}^N(f(\mathbf{x}_i)-y_i)^2$. 
    The final solution $\boldsymbol{\alpha}^*$ can be explicitly written as:
    \begin{equation}\label{alpha star}
    \boldsymbol{\alpha}^{\hat{\theta}_\mathcal{D}} = g_{T^*}(g_{T^*-1}(\cdots(g_1(\boldsymbol{\alpha}^{\nu^*,0}, \boldsymbol{\alpha}^{\nu^*,-1})))).
    \end{equation}
    and $g_1(\boldsymbol{\alpha}^{\nu^*,0}, \boldsymbol{\alpha}^{\nu^*,-1}) = \frac{\omega^{\nu^*,1}} {|{\mathcal{D}}|}{\mathbf{y}}$.
    Then, the key algorithm is $\mathscr{A}_{\hat{\theta}_\mathcal{D}}:\mathcal{D} \mapsto f(\mathbf{x})=\sum_{i=1}^{N}\boldsymbol{\alpha}^{\hat{\theta}_\mathcal{D}}_iK(\mathbf{x},\mathbf{x}_i)$.
    % In $ \nu $ method, let $ \widehat{\boldsymbol{K}} \in \mathbb{R}^{M \times M}$ denote the kernel matrix with entries $ \widehat{\boldsymbol{K}}_{ij} = K(\widehat{\boldsymbol{x}}_i,\widehat{\boldsymbol{x}}_j) $, and let $\widehat{\mathbf{y}}=[\widehat{y}_1,\widehat{y}_1,\cdots,\widehat{y}_M]^\top \in \mathbb{R}^M$ be the corresponding target vector, then $ \boldsymbol{\alpha} $ is iteratively computed updated (with initialization $ \boldsymbol{\alpha}^0 = \boldsymbol{\alpha}^{-1}  = \mathbf{0} $) via:
    % After $T$ iterations, the final parameters $\boldsymbol{\alpha}^* = \boldsymbol{\alpha}^{\hat{\theta}_\mathcal{D}}$ encapsulate the nonlinear relationships discovered in the original dataset.  $\hat{\theta}_\mathcal{D}=\{\nu_{\mathcal D},T_{\mathcal D}\}$. 
\end{itemize}

This knowledge encapsulation step is foundational. The  key algorithm, $\mathscr{A}_{\hat{\theta}_\mathcal{D}}$, captures the essential predictive relationships $\mathscr{A}_{\hat{\theta}_\mathcal{D}}(\mathcal{D})$ one wishes to share, providing the blueprint for the subsequent forging of a synthetic dataset with controllable, asymmetric utility.

\subsection{Asymmetric utility forging: overfitting synthesis data to the inference model}

% For synthetic data, buyers and sellers typically have differing objectives. Buyers prioritize the dataset’s predictive utility and efficiency, aiming to achieve strong predictive performance with minimal computational cost. Sellers, on the other hand, emphasize competitiveness, seeking to ensure that users cannot obtain high-quality predictions by introducing external data or employing unauthorized inference models. To simultaneously accommodate the needs of both buyers and sellers, 

Following the knowledge encapsulation stage where we defined the key algorithm $\mathscr{A}_{\hat{\theta}_\mathcal{D}}$, we now proceed to the second stage: {\it asymmetric utility forging}. Our goal here is to generate a synthetic dataset,  $ \mathcal{\widehat{D}}  = \{(\widehat{\mathbf{x}}_j, \widehat{y}_j)\}_{j=1}^M$,  that possesses two core properties of rivalry: model specificity and low compatibility. The central strategy is to craft a dataset $\mathcal{\widehat{D}}$ that intentionally  overfits  the  model (key model) produced by key algorithm.

To achieve this, we frame the data synthesis problem within a meta-learning paradigm. Instead of optimizing a model's parameters, we directly optimize the synthetic dataset $\mathcal{\widehat{D}}$ itself. The process is as follows: given a candidate synthetic dataset $\mathcal{\widehat{D}}$,  we first use it to get an inference model, $\widehat{f} = \mathscr{A}_{\hat{\theta}_\mathcal{D}}(\mathcal{\widehat{D}})$, by the key algorithm $\mathscr{A}_{\hat{\theta}_\mathcal{D}}$. We then evaluate this model's performance on the original  dataset, $\mathcal{D}$, which serves as a validation set. This step quantifies the utility of $\mathcal{\widehat{D}}$: a high-quality synthetic dataset should produce a model $\widehat{f}$ that accurately predicts the true outcomes in $\mathcal{D}$. Using a loss function $ \mathcal{L}(\cdot, \cdot) $  (e.g., squared loss for regression tasks), the predictive capability of the synthetic dataset can be quantified by the following validation loss:
\begin{equation} \label{loss}
	\frac{1}{N}\sum_{i=1}^{N} \mathcal{L}(\widehat{f}(\mathbf{x}_i), y_i)=\frac{1}{N}\sum_{i=1}^{N} \mathcal{L}\big((\mathscr{A}_{\hat{\theta}_\mathcal{D}}(\mathcal{\widehat{D}}))(\mathbf{x}_i), y_i\big).
\end{equation}
To generate the optimal rival synthetic dataset, we therefore formulate the optimization problem \eqref{target-111} that seeks the dataset $\widehat{\mathcal{D}}$ to minimize the validation loss, i.e.,
\begin{equation} \label{objective L}
	\mathcal{\widehat{D}} = \arg\min_{\mathcal{\widehat{D}},|\widehat{\mathcal{D}}|\leq M}\frac{1}{N}\sum_{i=1}^{N} \mathcal{L}(\widehat{f}(\mathbf{x}_i), y_i).
\end{equation}
% Note that for a fixed key algorithm $\mathscr{A}_{\hat{\theta}_\mathcal{D}}$, this validation loss is a well-defined function of the synthetic data points, making the optimization feasible.
The optimization formulation (\ref{objective L}) precisely illustrates how the two required properties of rivalry are engineered.
On one hand, model specificity is achieved by deliberately making the synthetic data  $\widehat{\mathcal{D}}$ to overfit the resulting model $\widehat{f}$, which is accomplished using two primary tactics: (1) we strictly limit the synthetic data volume $M$ (typically $M \ll N$) to ensure the model's parameter capacity exceeds the data complexity; and (2) we deliberately avoid any form of regularization in the optimization objective. 
This approach is counter-intuitive because it reframes overfitting from a liability into an asset. In conventional training, overfitting is detrimental because it makes a model dependent on a specific dataset, hindering generalization. Here, we invert this relationship: by making the dataset dependent on a specific model, overfitting becomes the very mechanism that forges the data's strategic value and ensures its utility is confined to the key algorithm. This turns a traditionally undesirable property into a strategic advantage, yielding a dataset with patterns that are maximally informative only when interpreted by the intended model.
On the other hand, low compatibility emerges as a natural byproduct of our optimization objective. The goal of (\ref{objective L}) is to maximize predictive utility, not to replicate the original  data distribution $P(\mathbf{x},y)$. As a result, the generated dataset $\widehat{\mathcal{D}}$ will have a distribution that diverges significantly from that of $\mathcal{D}$. This distributional shift ensures that simply merging $\widehat{\mathcal{D}}$ with other datasets hardly to improve the performance of model, and may even degrade.

Adopting the squared loss, $ \mathcal{L}(f(\mathbf{x}), y) = (f(\mathbf{x}) - y)^2 $, our final objective function for data synthesis becomes:
\begin{equation} \label{objective square loss}
	\mathcal{\widehat{D}} = \arg\min_{\mathcal{\widehat{D}},|\mathcal{\widehat{D}}|\leq M}\frac{1}{N}\sum_{i=1}^{N} \big(\widehat{f}(\mathbf{x}_i)- y_i\big)^2.
\end{equation}
The specific implementation of this objective depends on the structure of the key algorithm: 
\begin{itemize}
    \item \textbf{Models without hyperparameter ($\theta=\varnothing$)}: Taking linear regression with key algorithm $\mathscr{A}_{\hat{\theta}_\mathcal{D}}(\mathcal{\widehat{D}})=\widehat{f}({\mathbf{x}}) = (\mathbf{w}^*)^\top {\mathbf{x}} $ as an example, substituting the closed-form solution of $ \mathbf{w}^* $ into the objective (\ref{objective square loss}) yields the following optimization problem for generating a rival synthetic dataset:
    \begin{equation*} \label{loss linear regression}
    	\mathcal{\widehat{D}} = \arg\min_{\mathcal{\widehat{D}},|\widehat{\mathcal{D}}|\leq M}\frac{1}{N}\sum_{i=1}^{N} \bigg(\Big(\big(\widehat{\mathbf{X}}^\top \widehat{\mathbf{X}}\big)^{-1} \widehat{\mathbf{X}}^\top \widehat{\mathbf{y}}\Big)^\top\mathbf{x}_i - y_i\bigg)^2,
    \end{equation*}
    \item \textbf{Models with hyperparameter ($\theta\neq\varnothing$)}: Taking $\nu$-method with key algorithm $\mathscr{A}_{\hat{\theta}_\mathcal{D}}(\mathcal{\widehat{D}})=\widehat{f}({\mathbf{x}}) =  \sum_{j=1}^{M}\boldsymbol{\alpha}^{\hat{\theta}_\mathcal{D}}_jK(\mathbf{x}_i,\widehat{\mathbf{x}}_j)$ and the optimal hyperparameters $\hat{\theta}_\mathcal{D}=\{\nu^*,T^*\}$ as an example, substituting the iterative solution of $ \widehat{\boldsymbol{\alpha}}^{\hat{\theta}_\mathcal{D}} $ into the objective (\ref{objective square loss}) yields the following optimization problem for generating a rival synthetic dataset:
    \begin{equation} \label{loss nu method}
	\mathcal{\widehat{D}} = \arg\min_{\mathcal{\widehat{D}},|\widehat{\mathcal{D}}|\leq M}\frac{1}{N}\sum_{i=1}^{N} \bigg(\sum_{j=1}^{M}\widehat{\boldsymbol{\alpha}}^{\hat{\theta}_\mathcal{D}}_jK(\mathbf{x}_i,\widehat{\mathbf{x}}_j) - y_i\bigg)^2,
    \end{equation}
    where $\widehat{\boldsymbol{\alpha}}^{\hat{\theta}_\mathcal{D}} = g_{T^*}(g_{T^*-1}(\cdots(g_1(\widehat{\boldsymbol{\alpha}}^{\nu^*,0}, \widehat{\boldsymbol{\alpha}}^{\nu^*,-1}))))$ is fully determined by the synthetic dataset $\mathcal{\widehat{D}}$. 
\end{itemize}

% It is important to note that for a fixed training algorithm $\mathscr{A}$, the resulting inference model $\widehat{f} = \mathscr{A}(\mathcal{\widehat{D}})$ is solely determined by the synthetic dataset $\widehat{\mathcal{D}}$. Therefore, the validation loss (\ref{loss}) can be regarded as a function of the synthetic data samples.
% To obtain the most predictive synthetic dataset, we formulate an optimization problem that directly minimizes validation loss (\ref{loss}) with respect to $\widehat{\mathcal{D}}$:

% In this way, the task of generating synthetic data is reformulated as an optimization problem with respect to the synthetic dataset $ \widehat{\mathcal{D}} $.
% We adopt the square loss as the loss function, i.e., $ \mathcal{L}(f(\mathbf{x}), y) = (f(\mathbf{x}) - y)^2  $, and then (\ref{objective L}) can be written as
% \begin{equation} \label{objective square loss}
% 	\mathcal{\widehat{D}} = \arg\min_{\mathcal{\widehat{D}}}\frac{1}{N}\sum_{i=1}^{N} \big(\widehat{f}(\mathbf{x}_i)- y_i\big)^2,
% \end{equation}
% which is our objective function for data synthesis.

% Asymmetric Utility Forging. We construct $\widehat{\mathcal{D}}$ to overfit the ``key" model $\mathscr{A}_{\hat{\theta}_\mathcal{D}}(\cdot)$, that is $\mathcal{\widehat{\mathcal{D}}} = \arg\min\limits_{\mathcal{\widehat{\mathcal{D}}}}\frac{1}{N}\sum_{i=1}^{N} \mathcal{L}(\widehat{f}(\mathbf{x}_i), y_i)$, where $\widehat{f}(\cdot)= \mathscr{A}_{\hat{\theta}_\mathcal{D}}(\mathcal{\widehat{D}})$.

\subsection{Encapsulation-Forging (EnFo) Framework}

 {Based on the previous discussions, a typical realization of our Encapsulation-Forging (EnFo) framework is formalized in Algorithm \ref{algorithm}, in which we adopt  $\nu$-method as the inference model and solve the optimization problem (\ref{loss nu method}) via ADAM\endnote{ADAM is among the most widely used stochastic gradient descent (SGD) optimization algorithms. Its implementation details and relevance to our problem are discussed in the supplementary material (Appendix A).}, a representative stochastic gradient descent algorithm.} 

\vspace{0.15in}
\begin{breakablealgorithm}
    \small
	\caption{EnFo Framework (Using $\nu$-Method As Inference Model)}\label{algorithm}
	\begin{algorithmic}
		\Statex  \hspace{-0.15in} \textbf{Input:} A original  dataset $\mathcal{D}  = \{(\mathbf{x}_i, y_i)\}_{i=1}^N$, the number of samples in the synthetic dataset $M$, batch size $ B $ and the number of epochs $E$.
        
		\Statex \hspace{-0.05in} \(\triangleright\) \textbf{\textit{Stage 1  (Knowledge Encapsulation)}}
		\State  \hspace{-0.05in}  \textbf{1. Select the best hyperparameter of $\nu$-method on $\mathcal{D}$:}
        		\State   \hspace{0.15in} (1) Select hyperparameter the number of iterations $T$ by cross-validation: 
                $$T^* = \arg\min_{T} \frac{1}{|\mathcal{D}_{val}|} \sum_{i=1}^{|\mathcal{D}_{val}|} \bigg(\sum_{j=1}^{|D_{tr}|}   \boldsymbol{\alpha}^{\nu,T}_j K(\mathbf{x}_i,\mathbf{x}_j) - y_i\bigg)^2,$$ 
                \State   \hspace{0.15in} where $\mathcal{D}_{val}$ denotes the validation dataset and $D_{tr}$ denotes the training dataset.
        
		\Statex \hspace{-0.05in} \(\triangleright\) \textbf{\textit{Stage 2 (Asymmetric Utility Forging)}}
		\State  \hspace{-0.05in}  \textbf{2. Generate  the synthetic dataset (solve the optimization problem (\ref{loss nu method}) by ADAM):}
			\State    \hspace{0.15in} (1) Initialize $\widehat{\mathcal{D}} = \{(\widehat{\mathbf{x}}_j, \widehat{y}_j)\}_{j=1}^M$ by randomly sampling $M$ examples from $\mathcal{D}$.
            \State    \hspace{0.15in} (2) Use mini-batch SGD to train the model:
            \State    \hspace{0.15in} \textbf{for} $epoch \gets 1$ \textbf{to} $E$ \textbf{do}
        % \For{$epoch \gets 1$ \textbf{to} $E$}
            \State \hspace{0.35in} Divide $\mathcal{D}$ into mini-batches $\{\mathcal{B}_1, \mathcal{B}_2, \cdots, \mathcal{B}_K\}$ of size $B$;
                \State    \hspace{0.35in} \textbf{for} $ k \gets 1$ \textbf{to} $K$ \textbf{do}
                \State    \hspace{0.55in} Compute solution $\boldsymbol{\alpha}^{T^*}$ on the current $\mathcal{\widehat{D}}$: $ \boldsymbol{\alpha}^{T^*}=g_{T^*}(g_{T^*-1}(\cdots(g_1(\widehat{\boldsymbol{\alpha}}^{\nu^*,0}, \widehat{\boldsymbol{\alpha}}^{\nu^*,-1})))) $, where 
                \State    \hspace{0.75in} $g_1(\boldsymbol{\alpha}^{\nu^*,0}, \boldsymbol{\alpha}^{\nu^*,-1}) = \frac{\omega^{\nu^*,1}} {M}{\widehat{\mathbf{y}}}$, $\widehat{\mathbf{y}}=[\widehat{y}_1,\widehat{y}_2,\cdots,\widehat{y}_M]^\top$;
                \State \hspace{0.55in} Compute $k$-th batch loss $\mathcal{L}_k = \frac{1}{B}\sum_{i=1}^{B} \bigg(\sum_{j=1}^{M}\boldsymbol{\alpha}^{T^*}_jK(\mathbf{x}_i,\widehat{\mathbf{x}}_j) - y_i\bigg)^2,~~(\mathbf{x}_i, y_i)\in \mathcal{B}_k$; 
                \State \hspace{0.55in} Compute $\nabla \mathcal{L}_k$, the gradient of $\mathcal{L}_k$ with respect to the synthetic dataset $\widehat{\mathcal{D}}$;
                \State \hspace{0.55in} Update $\mathcal{\widehat{D}}$ via ADAM optimizer:
                $\mathcal{\widehat{D}} \leftarrow \text{Adam}(\mathcal{\widehat{D}}, \nabla\mathcal{L}_k)$;
            \State \hspace{0.35in} \textbf{end for}
        \State \hspace{0.15in} \textbf{end for}
		\Statex  \hspace{-0.15in} \textbf{Output:} A synthetic dataset $\mathcal{\widehat{D}} = \{(\widehat{\mathbf{x}}_j, \widehat{y}_j)\}_{j=1}^M$.
	\end{algorithmic}
\end{breakablealgorithm}
	\vspace{0.05in}
	\noindent\parbox{0.98\textwidth}{
		\footnotesize
		\textbf{Note:} For simplicity of training, we treat $\nu$ as a constant rather than a tunable hyperparameter, and set $\nu = 5$ following the original paper by \citet{gerfo2008spectral}.}
        \vspace{0.1in}

%讲讲算法的好处：在合成数据的过程中设计了效用，并且是专有效用，使得非竞争性的数据转变为竞争性的合成数据产品；讲数据量；
The EnFo framework is fundamentally distinguished from conventional generative models, such as GANs or DM. Rather than attempting to learn and replicate the original  data distribution, our objective in (\ref{objective square loss}) is engineered to directly maximize the utility of the synthetic data for a pre-specified task. This represents a paradigm shift: we embed strategic control over data utility directly into the synthesis process, a dimension not explicitly addressed by prior methods.

This utility-centric design yields several significant advantages for the data owner. First and foremost, it engineers rivalry into the synthetic data. By design, the resulting asset is highly valuable only when used as a standalone dataset with the pre-specified key algorithm. This provides the data owner with strict, technically-enforced control over how and by whom the shared data is used, mitigating the risks of unauthorized repurposing or value dilution. Second, the framework is highly efficient. Stemming from our strategic use of overfitting, it can produce a synthetic dataset where the data volume ($M$) is significantly smaller than the original ($N$) without sacrificing predictive utility. This translates to substantial practical benefits, including lower costs for data storage, transmission, and model training. Finally, the synthesis process provides inherent privacy protection, a property we empirically validate in the experimental section. Moreover, our framework is also trained using mini-batch SGD, which offers scalability and velocity advantages when handling large-scale or streaming data.

The EnFo framework detailed thus far, however, operates under a key scenario: the data recipient has a single, uniquely specified inference model and no existing data. This condition, while powerful for control, can be restrictive in scenarios where a partner has some but not certain models or wishes to augment their own data. Addressing this important extension is the focus of the next section, where we relax this constraint to broaden the applicability of our approach.

\section{Model Extensions} \label{sec: model extensions}
% The proposed base generative model is intentionally designed to produce synthetic data with strong competitiveness, such as exhibiting substantial statistical divergence from the original data, and achieving high predictive performance only when the inference model is trained under the designated algorithm. 
%This level of specificity plays a critical role in preventing unauthorized reuse or secondary commercial exploitation. 
%However, in certain application scenarios, users may prefer to relax such competitiveness in exchange for synthetic data with richer information content or broader applicability.
% To accommodate these diverse needs, it is essential for the generative framework to support flexible and controllable extensions. 
Building on the limitations identified previously, this section extends our framework to address more realistic and flexible marketing partnerships. We relax the strict single-model and standalone-use constraints to tackle two critical scenarios: first, creating data compatible with a partner's a set of inference model, and second, enabling data augmentation for partners who wish to enrich their own datasets. 
We demonstrate that both extensions can be elegantly achieved by incorporating strategic regularization terms into our core optimization objective.

% Our approach naturally lends itself to such extensibility, as the synthetic data are directly optimized as learnable parameters. 
% This design allows us to easily incorporate structural constraints or auxiliary objectives through regularization, enabling the model to adapt its behavior—for instance, by improving statistical fidelity or enhancing model adaptation to alternative inference models—according to user-specific requirements. 
% Below, we introduce two extended variants of the base model, aimed respectively at enhancing statistical alignment and improving model adaptability.

\subsection{Scenario 1: Data Augmentation}\label{section: Extended Model for Statistical Alignment}
Consider a common strategic challenge in marketing: a firm needs to build robust predictive models but possesses a dataset that is high-quality yet insufficient in volume. For example, a CPG company launching a new product in a niche market may have a small ``seed" dataset of early adopters. Similarly, a financial institution might have limited historical data for an emerging customer segment. In these situations, the goal is not to add new features (i.e., expand data attributes), but to augment the existing data by generating more samples that share the same statistical characteristics. This requires a fundamental shift in our synthesis objective: instead of creating a standalone-use dataset, we must now synthesis data designed to be synergistically combined with an existing dataset.

To achieve this, we introduce a regularization term to our original optimization objective (\ref{objective square loss}). This term explicitly enforces statistical alignment between the synthetic data $\mathcal{\widehat{D}}$ and the original data $\mathcal{D}$. The new objective function is:
% To enhance the statistical alignment between the synthetic and original data distributions, we augment the original optimization objective (\ref{objective square loss}) by introducing a regularization term based on the measurement of distributed distance as follows:
$$
    \mathcal{\widehat{D}} = \arg\min_{\mathcal{\widehat{D}}}\frac{1}{N}\sum_{i=1}^{N} \big(\widehat{f}(\mathbf{x}_i), y_i\big)^2 + \lambda\cdot dis(\mathcal{\widehat{D}}, \mathcal{D}),
$$
where $ \lambda > 0 $ is a hyperparameter controlling the tradeoff between predictive fidelity (the first term) and the statistical similarity between the two datasets (the second term). The function $dis(\cdot, \cdot)$ is a measure of the distance between the empirical distributions of the synthetic and original data. 
A representative choice for this distance metric is the Wasserstein distance, denoted by $ W(\mathcal{\widehat{D}}, \mathcal{D}) $, which leads to the following specific optimization problem:
%denotes the Wasserstein distance between the empirical distributions of the synthetic and original datasets.
%the Wasserstein distance as follows:
\begin{equation} \label{objective Wasserstein}
	\mathcal{\widehat{D}} = \arg\min_{\mathcal{\widehat{D}}}\frac{1}{N}\sum_{i=1}^{N} \big(\widehat{f}(\mathbf{x}_i), y_i\big)^2 + \lambda\cdot W(\mathcal{\widehat{D}}, \mathcal{D}),
\end{equation}
It is worth noting that the Wasserstein distance can be efficiently and stably approximated in PyTorch using the Geometric Losses package, which provides differentiable formulations suitable for training via gradient descent. By minimizing this combined objective function  (\ref{objective Wasserstein}), we generate a synthetic dataset that not only produces an accurate predictive model but also faithfully preserves the statistical characteristics of the original data,  making it ideal for augmentation.

\subsection{Scenario 2: Inference Model Adaptability}\label{section: Extended Model for Inference Model Adaptability}
Beyond augmenting data volume, a crucial challenge in data partnerships is accommodating a partner's diverse analytical workflows. For example, when a firm shares data with a marketing agency, the agency may not rely on a single bespoke model. Instead, it often employs a standard  ``analytics toolkit"—a portfolio of models like logistic regression, kernel ridge regression, and support vector machine—to serve various client needs. In this common scenario, the synthetic data must be versatile and robust, delivering high predictive utility not for one specific model  but across the entire pre-approved set. 

To achieve this, we introduce a second extension aimed at improving the model adaptability of the synthetic data $\mathcal{\widehat{D}}$.  The key idea is to control the complexity of the synthetic dataset, which directly influences the training error of inference models—simpler datasets are typically easier to fit and thus result in lower fitting errors. Taking that into account, we introduce a regularization term  $ \frac{1}{M}\sum_{j=1}^{M} \big(\widehat{f}(\widehat{\mathbf{x}}_j ) - \widehat{y}_j\big)^2 $ to directly control the training error of the inference model on the synthetic data $ \mathcal{\widehat{D}}= \{(\widehat{\boldsymbol{x}}_j, \widehat{y}_j)\}_{j=1}^M $. In this case, the extended model has the following form:
\begin{equation} \label{objective final}
	\mathcal{\widehat{D}} = \arg\min_{\mathcal{\widehat{D}}}\frac{1}{N}\sum_{i=1}^{N} \big(\widehat{f}(\mathbf{x}_i) - y_i\big)^2 +\beta\cdot \frac{1}{M}\sum_{j=1}^{M} \big(\widehat{f}(\widehat{\mathbf{x}}_j) - \widehat{y}_j\big)^2,
\end{equation}
where $ \beta $ is the trade-off parameter.  By encouraging the synthetic data to be easier to fit, this objective improves their compatibility with alternative inference models, thus supporting better adaptability beyond the designated training architecture.

The above two extensions illustrate how our framework can flexibly accommodate different objectives through targeted regularization. Beyond these examples, the proposed method readily supports the incorporation of a wide range of structural constraints into the synthetic data. This highlights the flexibility and controllability of the proposed framework in adapting to diverse user requirements and application contexts.

\section{Metrics and Benchmarks}\label{sec:Methodology}

In this section, we present a comprehensive evaluation framework for assessing our proposed data synthesis approach against benchmark methods. This evaluation is anchored in five principal dimensions: \textit{rivalry}, \textit{data volume}, \textit{utility}, as well as its extensibility in terms of  \textit{data augmentation} and \textit{inference model adaptability}.

\subsection{Rivalry}
Rivalry reflects the extent to which synthetic data limits the potential for unintended secondary use—ensuring that it supports specific modeling objectives while remaining ineffective for broader misuse such as repurposing, reverse engineering, or commercial exploitation. As defined in Section~\ref{Problem Setting}, a synthetic dataset exhibiting rivalry should simultaneously satisfy low compatibility and model specificity. Accordingly, we evaluate rivalry from two key aspects: the external data and the inference model. In addition, since privacy protection is a fundamental objective of data synthesis, we incorporate it as a critical metric within this dimension as well.

\subsubsection{Rivalry with Respect to External Data}
Ensuring low compatibility with external datasets is essential, as it prevents the third party users from augmenting the synthetic data with their own proprietary datasets to extract additional value or repurpose it. This property is typically evidenced by degraded predictive performance when the synthetic dataset is combined with external data.
To achieve this, the synthetic data should exhibit substantial statistical divergence from the real data.
Accordingly, we evaluate the rivalry with respect to external data using two metrics:

$\bullet$ \textbf{Statistical Divergence: } Statistical divergence captures the extent to which synthetic data intentionally deviates from the statistical structure of the original data, thereby ensuring low compatibility and preventing effective combination with user's own datasets. To quantify this divergence, we compute statistical distances between the distributions of synthetic and original datasets. Specifically, we employ three measures widely used  in the statistics and
marketing literature: Kullback–Leibler (KL) divergence \citep{kullback1951information}, Jensen–Shannon divergence (JSD) \citep{lin2002divergence}, and Wasserstein distance \citep{kantorovich1939mathematical}. All three measures capture discrepancies in the underlying probability distributions, with larger values indicating greater deviation. The definitions and calculations of the three statistical measures are detailed in the supplementary material (Appendix B).

$\bullet$ \textbf{Resistance to External Augmentation: } Resistance to external augmentation reflects the extent to which the predictive performance of synthetic data degrades when additional user-provided samples are incorporated. To evaluate this, we augment the synthetic dataset with a fixed number of user-owned observations and retrain the inference model. We then compare the predictive performance before and after augmentation. A significant drop in predictive accuracy indicates strong resistance to external augmentation and higher rivalry, thereby limiting users’ potential for secondary development and commercial exploitation.

Notably, since synthetic data is primarily used to train regression models for predicting new observations, we employ the root mean squared error (RMSE), one of the most commonly used evaluation metrics in marketing, economics, and related fields \citep{evgeniou2007convex, huang2016consumer, ansari2018probabilistic, wei2025estimating}, to evaluate the predictive performance of inference models on the test set. Formally, for $ M  $ number of test observations $  \{(\mathbf{x}_j, y_j)\}_{j=1}^M $, the RMSE of the inference model $f$ is computed as:
$
	RMSE = \sqrt{\frac{1}{M}\sum_{j=1}^{M}\big(y_j - \hat{y}_j\big)^2},
$
% \begin{equation}\label{RMSE}
% 	RMSE = \sqrt{\frac{1}{M}\sum_{j=1}^{M}\big(y_j - \hat{y}_j\big)^2},
% \end{equation}
where $ y_j $ denotes the true outcome of the $j$-th test observation, and $ \hat{y}_j = f(\mathbf{x}_j) $ is the corresponding predicted output. A lower RMSE indicates more accurate predictions, thereby reflecting a stronger predictive capability of the synthetic data.

\subsubsection{Rivalry with Respect to Inference Model}
We employ the inference model specificity to measure the rivalry of synthetic data with respect to the inference model, which reflects the extent to which synthetic data is intentionally restricted to a pre-specified inference model, with performance degrading significantly when transferred to other models.
To evaluate this, we train inference models on the synthetic dataset using both the pre-specified model and two widely used alternatives: kernel ridge regression (KRR) and support vector regression (SVR). We then compare their predictive performances on a common test set. A significant performance gap—where the pre-specified model achieves strong results while the alternatives perform poorly—indicates high model specificity. 
This implies that the synthetic data cannot be easily repurposed for modeling tasks beyond the scope of the intended architecture, thereby reducing the risk of secondary exploitation or unauthorized misuse.

\subsubsection{Privacy Protection}
Privacy reflects the extent to which synthetic data mitigates the risk of re-identification and protects sensitive individual information from being exposed.
We employ Monte Carlo Attack Accuracy (\textit{MCAA}) as the evaluation metric for data privacy protection \citep{benjamin2019monte}. The Monte Carlo attack is a representative membership inference attack (MIA) technique \citep{shokri2017membership, zhang2021membership} against generative models, aiming to identify whether a specific individual belongs to the training data of the generative model. 
% MIAs pose a concrete threat to data privacy, particularly in applications involving individual-level behavioral or transactional data. If an adversary can determine that a specific individual’s data was used to train the generative model, they may infer sensitive characteristics—such as medical conditions, purchasing habits, or personal preferences—especially when attackers have access to auxiliary data for linkage. 
% Therefore, the success rate of MIAs serves as a practical indicator of the privacy protection capability. 
We adopt the accuracy of Monte Carlo attacks as our evaluation metric, as it relies solely on the similarity between observations and makes no assumptions about the underlying generative models. This makes the approach both simple to implement and applicable to a wide range of model architectures. A lower \textit{MCAA} indicates that the synthetic data is less susceptible to membership inference attacks, thereby offering stronger privacy protection. Details of the definition and calculation of \textit{MCAA} are provided in the supplementary material (Appendix C).

It is important to note that when evaluating privacy protection, one must also consider its trade-off with data utility.
Actually, the tradeoff between data utility and privacy protection has been a longstanding concern in the data sharing and disclosure literature. As formalized in the risk–utility framework by \citep{duncan2004disclosure}, efforts to reduce disclosure risk often come at the expense of the utility of the data for downstream analytical tasks. 
Traditional masking techniques—such as coarsening, top-coding, and noise injection—aim to reduce re-identification risk but often degrade data utility. Synthetic data generated by modern generative models (e.g., GANs, diffusion models) has emerged as a promising alternative, offering a better balance between accessibility and privacy. However, such models may memorize training data, exposing them to privacy risks like membership inference attacks \citep{mukherjee2021privgan}. As a result, the fundamental trade-off between privacy protection and data utility remains a central concern \citep{anand2023using}.
% Traditional masking strategies—such as coarsening, top-coding, or adding statistical noise—aim to limit identifiability by distorting sensitive information, thereby reducing the risk of privacy disclosure. However, the added perturbations or noise typically degrade the quality of inference, leading to a substantial loss in data utility. Recent years have seen a surge of interest in using synthetic data generated by generative models, notably GANs and diffusion models, as a privacy-preserving approach to data sharing. While synthetic data is often viewed as a promising solution for balancing data accessibility and individual privacy, recent research has shown that high-capability generative models are prone to memorizing their training data—making them vulnerable to privacy leakage, especially through membership inference attacks \citep{mukherjee2021privgan}. Thus, the inherent tradeoff
% between disclosure risk and data utility remains an essential consideration in evaluating synthetic data approaches \citep{anand2023using}.

To formally evaluate this tradeoff, similar to \citep{schneider2018flexible, anand2023using}, we construct a two-dimensional scatter plot that visualizes the predictive capability (measured by RMSE) against Monte Carlo Attack Accuracy (MCAA) for various methods, where the two axes correspond to the utility of the data and the risk of disclosure, respectively. Obviously, an ideal synthetic data method should lie in the bottom left of the plot, which corresponds to low RMSE and low MCAA, reflecting both strong predictive capability and effective privacy protection.

\subsection{Data Volume}
Data volume refers to the extent to which the synthetic data is compact and efficient for transfer, storage, and model training, without compromising downstream predictive performance.
We measure data volume by its \textit{sample size}, i.e., the number of observations contained in the synthetic dataset. On one hand, in a standard dataset where each observation has a fixed storage size, the total number of observations directly determines the overall dataset size, which in turn affects the cost and efficiency of data storage and transfer—particularly in systems with bandwidth or memory constraints. On the other hand, for downstream inference models trained on synthetic data, the number of observations directly determines the computational cost, memory requirements, and training time, which, in turn, influence the efficiency and accessibility for users. Therefore, the number of observations serves as a critical metric of data volume when evaluating the practicality and scalability of synthetic data in real-world deployment scenarios—particularly in settings where efficiency, speed, and resource constraints are of concern.

It is important to note that a clear tradeoff also exists between predictive performance and data volume, forming a fundamental tension in data synthesis and downstream deployment. In many practical scenarios—particularly those involving data transfer or limited storage and computing resources—maintaining a smaller size of dataset is often desirable, as it can significantly improve transfer speed, reduce storage requirements, and lower the computational cost of model training. However, smaller data volumes often come at the expense of predictive performance, as fewer observations limit the amount of predictive information available to inference models. 
This inherent tradeoff motivates our analysis of sample efficiency: the extent to which the synthetic dataset can preserve sufficient predictive information even with significantly fewer observations. A sample-efficient dataset maintains high predictive capability under constrained data volume, effectively balancing utility and operational efficiency. To quantify this tradeoff, we evaluate the RMSE of the inference models trained on synthetic datasets  of varying sizes, and examine how predictive accuracy degrades as sample size decreases.

\subsection{Utility}
Utility reflects the training efficiency and stability of the generative model, which are critical for ensuring reliable and efficient deployment in real-world practical applications.  It is evaluated from three perspectives: \textit{stability}, \textit{scalability}, and \textit{velocity}, reflecting the generation consistency, computational efficiency with larger datasets, and adaptability to new data streams, respectively. We detail the interpretation and evaluation method of each below.

\subsubsection{Stability}
Stability captures the variability in predictive performance across multiple runs of data synthesis. It reflects how consistently a generative model can reproduce data that supports reliable inference. Stability is an important consideration in practical settings, as highly unstable generators may lead to unpredictable model performance and reduced trust in synthetic data. We quantify stability by computing the standard deviation of RMSE across multiple synthetic datasets. Formally, for $ K $ independently generated synthetic datasets, we compute the standard deviation of their RMSEs as:
$
	Std = \sqrt{\frac{1}{K} \sum_{k=1}^{K} \left(RMSE_k - \frac{1}{K} \sum_{k=1}^{K} RMSE_k\right)^2},
$
% \begin{equation}\label{stability}
% 	Std = \sqrt{\frac{1}{K} \sum_{k=1}^{K} \left(RMSE_k - \frac{1}{K} \sum_{k=1}^{K} RMSE_k\right)^2},
% \end{equation}
where $ RMSE_k $ denotes the predictive error of the $ k $-th synthetic dataset. A lower standard deviation indicates more stable performance and, therefore, a more reliable generative approach.

\subsubsection{Scalability}
Scalability reflects the extent to which the model maintains efficient training speed as the size of the training dataset increases. 
We evaluate the scalability of our framework based on its training time on datasets with different number of observations. Since our model is trained using stochastic gradient descent (SGD)-based methods, the choice of batch size can significantly affect the training time. To ensure comparability, we train the model to convergence on datasets of varying sizes using a fixed batch size. We record the per-iteration training time to evaluate how training cost scales with data volume.

\subsubsection{Velocity}
Velocity reflects the efficiency of the generative model in accommodating new observations, particularly in dynamic environments where new  observations continually arrive in streaming manner. Rather than retraining the model from scratch, a more efficient generator should support incremental updates with minimal computational overhead. To evaluate this, we follow the procedure used in  \citep{anand2023using}, and compare the performance and efficiency of two updating strategies: (1) ``streamed training'', i.e., new data is streamed into the previously trained model, without retraining from scratch; (2) ``restarted training'', i.e., the model is retrained from scratch on the combination of new and original training data. We begin by training the generative model to convergence on the original dataset. Then, we simulate a burst of new data and apply both update strategies. Velocity is assessed by comparing how quickly each approach improves predictive performance ($ RMSE $) after incorporating the new data. A faster improvement implies higher velocity and better adaptability in dynamic environments.

\subsection{Extensibility in Terms of Data Augmentation}
In Section \ref{section: Extended Model for Statistical Alignment}, we propose an extended model that relaxes the strict standalone-use constraint by explicitly enforcing statistical alignment between the synthetic and original data. The resulting synthetic dataset faithfully preserves the statistical characteristics of the original data, enabling users to safely augment their own data with synthetic samples and improve the accuracy of downstream inference models. Notably, the addition of synthetic data generated by the base model offers no performance benefit, owing to its substantial misalignment with the distribution of real data.

To evaluate the effect of data augmentation, we consider a downstream inference task under four training scenarios: (1)	using only the user’s own data;
(2)	user’s own data + synthetic data from the base model;
(3)	user’s own data + synthetic data from an extended model;
(4) user’s own data + the original  data used to train the generator.
A desirable pattern is that adding synthetic data from the base model degrades performance due to its distributional divergence, whereas adding data from the extended model improves performance, approaching the original  data case. This suggests that the extended model allows controllable statistical alignment that enables synthetic data to be safely and effectively used for data augmentation.
	
\subsection{Extensibility in Terms of Inference Model Adaptability}
In addition to the data augmentation, in Section \ref{section: Extended Model for Inference Model Adaptability} we introduce another extended model that relaxes the single-model constraint by preserving the predictive performance of the designated model even when applied to other commonly used inference models—referred to as inference model adaptability. To evaluate the model adaptability of the extended model (\ref{objective final}), we compare the predictive performance of synthetic data generated by the base rivalry model and its extended variant when used to train alternative inference architectures—specifically, kernel ridge regression (KRR) and support vector regression (SVR). The base model is expected to yield higher RMSE under KRR and SVR due to its rivalry design, particularly its emphasis on model specificity.  In contrast, the extended model is expected to enhance the adaptability of synthetic data, achieving comparable RMSE under these alternative inference models. This demonstrates the capability of the proposed generative model to balance rivalry and generalizability through controlled extensions.

\subsection{Benchmarks}
We employ data generation and privacy protection methods commonly used in the literature as benchmarks to comprehensively evaluate our proposed approach. A summary of the benchmark methods and their brief descriptions is presented in Table \ref{tab:benchmark-methods}.
\begin{table}[htbp]
	\centering
	\small
	\caption{Description of Benchmark Methods}
	\begin{tabular}{clp{10cm}}
		\toprule
		\# & Benchmark Method & Description \\
		\midrule
		1 & original  data & Original data without any protection. \\
        2 & original  data (Sampled) & A subset of the original dataset obtained by random sampling. \\
		3 & GAN & Synthetic data produced by generative adversarial network.\\
		4 & Diffusion & Synthetic data produced by diffusion model.\\
		5 & Random noise & Random noise is added to each attribute of the data. \\
		6 & Rounding & Each attribute is rounded to the nearest value at the second most significant digit. \\
		7 & Top coding & Each attribute is truncated at its 95th percentile. \\
		8 & 20\% (50\%) swapping & A random 20\% (50\%) of observations are divided into two groups, and all their attributes are swapped between the groups. \\
		\bottomrule
	\end{tabular}
	\label{tab:benchmark-methods}
\end{table}

\section{Empirical Context and Results}\label{sec:Experiments}
In this section, we empirically evaluate the proposed Encapsulation-Forging (EnFo) framework on both Monte Carlo and real-world datasets. For each scenario, we comprehensively assess the performance of our method relative to benchmark approaches across various dimensions.  In these evaluations, we adopt Algorithm~\ref{algorithm} as the representative implementation of our framework, with $\nu$ method serving as the inference model.
Specifically, following the evaluation framework in Section \ref{sec:Methodology}, we examine five principal dimensions: \textit{rivalry}, which reflects the synthetic data’s low compatibility with external data, its specificity to designated inference models, and its capability for privacy protection; \textit{data volume}, which reflects the sample efficiency, i.e., the tradeoff between predictive performance and data size; \textit{utility}, which demonstrates the  training efficiency and stability of the generative process; and the extensibility in terms of \textit{data augmentation} and \textit{model adaptability}, which reflect the extended models' capability to augment user’s own data effectively and generalize to a broader range of inference models, respectively. 
\subsection{Monte Carlo  Experiments}
\subsubsection{Datasets}
We first conduct empirical evaluations on the following two Monte Carlo simulated datasets. The details of each dataset are presented in the supplementary material (Appendix D).

$\bullet$ \textbf{CLV Prediction Data: } 
We simulate customer-level purchase data using the covariate-extended Pareto/NBD model (Abe model) \citep{schmittlein1987counting, abe2009counting}, a well-established framework for modeling repeat transactions and customer lifetime value (CLV). 
We simulate a cohort of $ 15,000 $ customers with $10$ randomly generated covariates representing individual-level characteristics, and track their purchase counts over a 32-week calibration period followed by a 32-week holdout period.
We split the data into $10,000$ training and $5,000$ test samples. The resulting task is to predict future purchase counts based on covariates in a regression setting.

$\bullet$ \textbf{Friedman Synthetic Regression Data: } 
To demonstrate the general applicability of our framework beyond marketing-specific scenarios, we construct a second dataset based on the well-known Friedman \#3 benchmark function \citep{friedman1991multivariate}, a standard non-linear regression simulation, with four input variables drawn from independent uniform distributions. The output is a nonlinear function of these inputs with added Gaussian noise. A total of $10,000$ samples are generated and split evenly into training and test sets. The task is framed as a standard regression problem.

\subsubsection{Rivalry} We evaluate the rivalry of our synthetic data across three aspects: rivalry with respect to external data, rivalry with respect to inference model, and privacy protection.  The corresponding experimental results and analyses are presented below.

\begin{figure}
	\FIGURE
	{\includegraphics[width=0.8\linewidth]{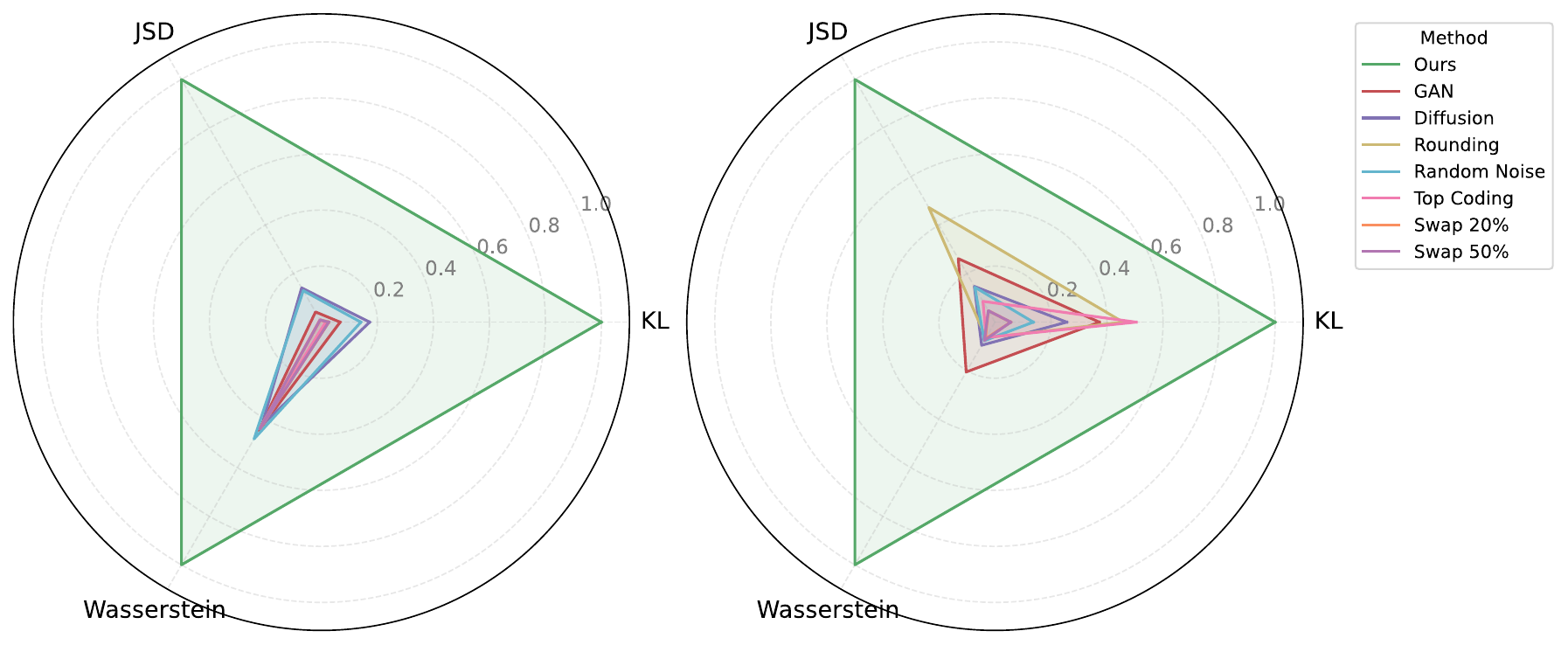}} 
	{Statistical Divergence Results on Monte Carlo Datasets (Higher Is Better) \label{figure MonteCarlo Statistics-Divergence}}
	{This radar chart reports the KL divergence, JSD  and Wasserstein distance between the original dataset and the synthetic datasets generated by our base model and benchmark methods on two Monte Carlo datasets. Left: CLV Prediction Data; Right: Friedman Synthetic Regression Data.}
\end{figure}

$\bullet$ \textbf{Rivalry with Respect to External Data: }  we evaluate the rivalry with respect to external data using two metrics: statistical divergence and resistance to external augmentation.
The statistical divergence results on the two Monte Carlo datasets are presented in Figure \ref{figure MonteCarlo Statistics-Divergence}, visualized using radar charts of the three distributional metrics: KL divergence, JSD, and Wasserstein distance. For visualization purposes, all three metrics in the radar charts are normalized by their respective maximum values to ensure consistency in scale. It can be observed that across both datasets, our base model consistently yields significantly higher values across all three metrics compared to the benchmark methods, indicating stronger statistical divergence from the original data. In terms of numerical values, on the CLV Prediction dataset, the highest KL divergence among the benchmark methods is 0.102 achieved by Diffusion, whereas our base model attains a substantially higher value of 0.590, which is nearly 6 times higher. Similarly, among the benchmark methods, the highest JSD and Wasserstein distance are 0.024 and 20.810, achieved by Diffusion and Random Noise, respectively. In contrast, our base model records significantly larger values of 0.171 and 43.239—approximately 7 times and 2 times greater, respectively. These findings are consistent with the results on the Friedman Synthetic Regression dataset. This superior statistical divergence enables our base rivalry model to generate synthetic data that statistically deviate from the original data distribution, thereby preventing effective data combination with external sources.

\begin{table}[h]
	\TABLE
	{Resistance to External Augmentation Results on Monte Carlo Datasets\label{tab: External Augmentation}}
	{	\begin{tabular}{ccccc}
			%\hline
			\toprule
			\vspace{-2.5mm}
			& \multicolumn{2}{c}{CLV} & \multicolumn{2}{c}{Friedman}\\ 
			
			\multirow{2}{*}{Sample size}& \hspace{0.05cm}\hrulefill\hspace{-0.6cm} & \hspace{-0.2cm}\hrulefill\hspace{0.05cm} &\hspace{0.05cm}\hrulefill\hspace{-0.6cm} & \hspace{-0.2cm}\hrulefill\hspace{0.05cm} \\
			& Synthetic data & Augmented data & Synthetic data & Augmented data		
			\\\hline
			$ 100 $  & 7.167 & 7.595 &  0.211 &  0.279
			\\
			$ 500 $  & 7.145 & 7.342 & 0.205  &  0.253
			\\
			$ 1000 $  & 7.142 & 7.239  &  0.204 &  0.242
			\\ \hline
			\toprule
	\end{tabular}}{This table reports the RMSE of our designated inference model under two settings: (1) trained on synthetic data generated by our base competitive model; and (2) trained on augmented data, which combines the synthetic data with additional user-owned samples.}
\end{table}

To assess the resistance of our synthetic data to external augmentation, we compare the predictive accuracy of inference models trained on two datasets: (1) the original synthetic dataset generated by our base rivalry model, and (2) an augmented dataset that combines the synthetic data with 1,000 additional user-owned samples.
The results for the two Monte Carlo datasets across sample sizes of 100, 500, and 1,000 are reported in Table \ref{tab: External Augmentation}. We observe that across both datasets, inference models trained on the augmented datasets consistently yield higher RMSEs than those trained on the original synthetic data, indicating that such external augmentation can degrade predictive performance. For example, at a sample size of 100, augmenting the dataset with external samples increases the RMSE from 7.167 to 7.595 on the CLV Prediction dataset, and from 0.211 to 0.279 on the Friedman Synthetic Regression dataset, corresponding to 6\% and 32\% increases, respectively. These results suggest that our synthetic data exhibits strong resistance to external augmentation, thereby reducing the risk of secondary development and unauthorized commercial exploitation.

$\bullet$ \textbf{Rivalry with Respect to Inference Model: }we assess the inference model specificity by comparing the predictive performance
of KRR and SVR with the designated model. The results with sample sizes of 100, 500, and 1000 are presented in Table \ref{tab: Model Specificity}, where Synthetic data (Base model) refers to the results trained on the synthetic data of our base rivalry model. Using the RMSE of the designated model as the reference, we observe that both KRR and SVR exhibit substantially higher RMSE when trained on Synthetic data (Base model). For instance, on the CLV Prediction dataset with a sample size of 100, the designated model achieves an RMSE of 7.083, while KRR and SVR yield substantially higher RMSEs of 8.171 and 8.058. These results demonstrate that our base rivalry model effectively restricts usability to the designated inference model, with non-designated models suffering substantial drops in predictive performance.

\begin{table}[h]
	\TABLE
	{Inference Model Specificity and Adaptability Capability Results on Monte Carlo Datasets\label{tab: Model Specificity}}
	{	\begin{tabular}{ccccccc}
			%\hline
			\toprule
			\vspace{-2.5mm}
			& & & \multicolumn{2}{c}{Synthetic data (Base model)} & \multicolumn{2}{c}{Synthetic data (Extended model)}\\ 
			
			\multirow{2}{*}{Dataset}&\multirow{2}{*}{Sample size}&\multirow{2}{*}{Designated model}& \hspace{0.05cm}\hrulefill\hspace{-0.6cm} & \hspace{-0.2cm}\hrulefill\hspace{0.05cm} &\hspace{0.05cm}\hrulefill\hspace{-0.6cm} & \hspace{-0.2cm}\hrulefill\hspace{0.05cm} \\
			&  &  & ~~~~~~~~KRR & SVR& ~~~~~~~~~~~KRR & SVR		
			\\\hline
			\multirow{3}{*}{CLV}    & $ 100 $ & 7.083 & ~~~~~~~~8.171  & 8.058&  ~~~~~~~~~~~7.124 &  7.138
			\\
			& $ 500 $ & 7.072 &  ~~~~~~~~7.595 &  7.849&  ~~~~~~~~~~~7.036&  7.078
			\\
			& $ 1000 $ & 7.016 & ~~~~~~~~7.401  &  7.488& ~~~~~~~~~~~7.038  &  7.071
			\\\cline{1-7}
			\multirow{3}{*}{Friedman}  & $ 100 $ & 0.215 &  ~~~~~~~~0.391 & 0.343&  ~~~~~~~~~~~0.273  &  0.311
			\\
			& $ 500 $ & 0.200 &  ~~~~~~~~0.256 &  0.351 & ~~~~~~~~~~~0.239 &  0.267
			\\
			& $ 1000 $ & 0.199 & ~~~~~~~~0.263  &  0.350 & ~~~~~~~~~~~0.255 &   0.266 
			\\ \hline
			\toprule
	\end{tabular}}{This table reports the RMSE of our designated inference model, compared against two alternative scenarios: (1) KRR and SVR trained on synthetic data generated by our base rivalcompetitive model; and (2) KRR and SVR trained on synthetic data generated by our extended model for inference model adaptability (see extended model  (\ref{objective final})).}
\end{table}

\begin{figure}[h]
	\FIGURE
	{\includegraphics[width=0.8\linewidth]{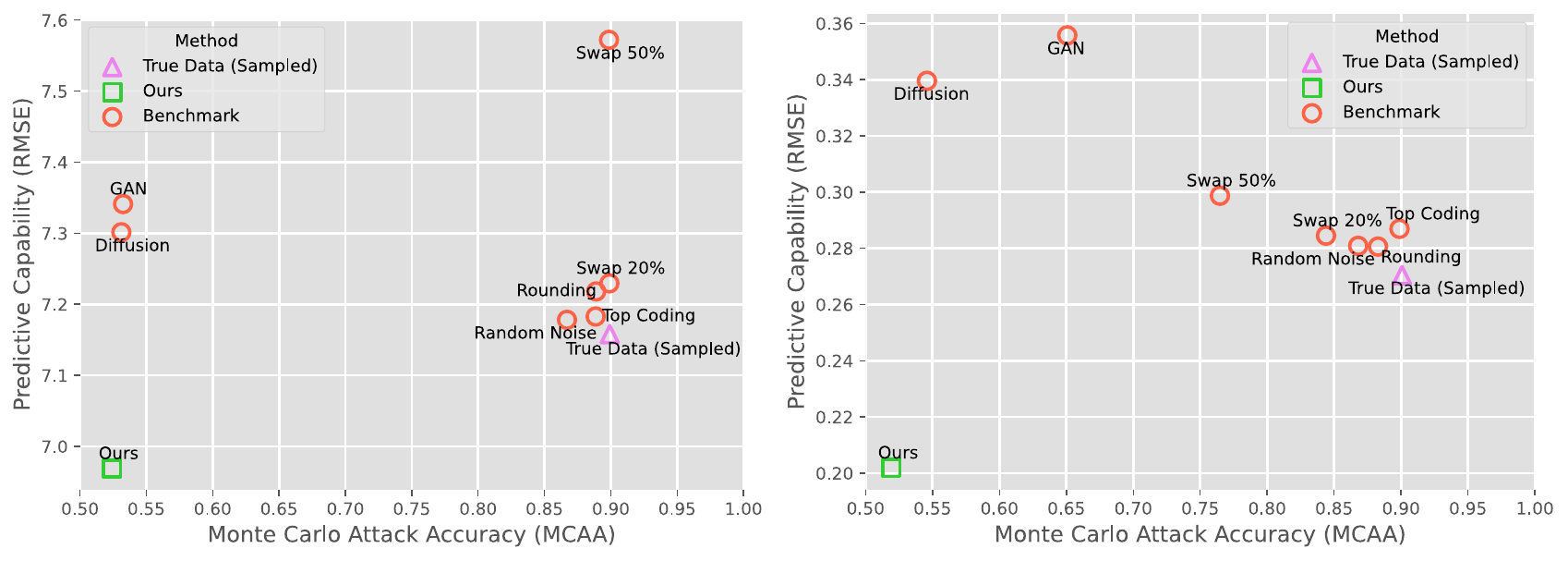}} 
	{Accuracy-Privacy Results on Monte Carlo Datasets \label{figure MonteCarlo Accuracy-Privacy}}
	{This figure reports the predictive capability (RMSE) and Monte Carlo Attack Accuracy (MCAA) for our framework and benchmark methods on the two Monte Carlo datasets, with a sample size of 3,000. A lower MCAA indicates stronger privacy protection. Left: CLV Prediction Data; Right: Friedman Synthetic Regression Data.}
\end{figure}

$\bullet$ \textbf{Privacy Protection: } 
We evaluate the balance between predictive accuracy (measured by RMSE) and privacy protection (measured by MCAA) for our proposed framework and benchmark methods. A lower RMSE indicates stronger predictive performance, while a lower MCAA reflects greater resistance to privacy attacks. Figure~\ref{figure MonteCarlo Accuracy-Privacy} reports the results on both Monte Carlo datasets, using a fixed sample size of 3,000. 

On the CLV Prediction dataset (left panel), the four benchmark methods—Rounding, Topcoding, Random Noise, and 20\% Swapping—achieve RMSE and MCAA values close to those of the original  data (Sampled), indicating limited privacy protection but minimal distortion to predictive accuracy. In contrast, 50\% Swapping yields a substantially higher RMSE than 20\% Swapping, due to the greater distortion caused by the increased swap ratio. The two generative models, GAN and Diffusion, achieve notably low MCAA values, thanks to their data synthesis mechanisms that enhance privacy by avoiding direct perturbation of the original data. However, this privacy advantage comes with a tradeoff in predictive accuracy, as both methods exhibit higher RMSEs compared to the aforementioned perturbation-based approaches. 

Obviously, an ideal synthetic data should achieve both high predictive accuracy and strong privacy protection, positioning itself in the bottom left of the RMSE-MCAA plot. In this regard, our proposed framework demonstrates a consistently favorable balance, exhibiting both lower RMSE and reduced MCAA compared to all benchmark methods. Specifically, our framework achieves an RMSE of 6.969, notably lower than the best benchmark result of 7.178 obtained from the original  data (Sampled), highlighting its superior predictive performance and sample efficiency. In terms of privacy protection, our framework achieves the lowest MCAA of 0.524, outperforming both generative models—GAN (0.532) and Diffusion (0.531)—and notably superior to the best perturbation-based method, Random Noise (0.867). It is worth noting that although the MCAA values of GAN and Diffusion are close to ours, their predictive performance is notably worse: GAN and Diffusion yield RMSEs of 7.341 and 7.301, respectively, which are significantly higher than the RMSE of 6.969 achieved by our framework. These finds are also the case for the Friedman Synthetic Regression dataset (right panel),
% where our framework achieves an RMSE of 0.202 and a MCAA of 0.519, both outperforming all benchmark methods. 
These results demonstrate that our framework consistently strikes a superior balance between accuracy and privacy to the benchmark methods across both datasets.

\subsubsection{Data Volume} 
\begin{figure}[h]
	\FIGURE
	{\includegraphics[width=0.8\linewidth]{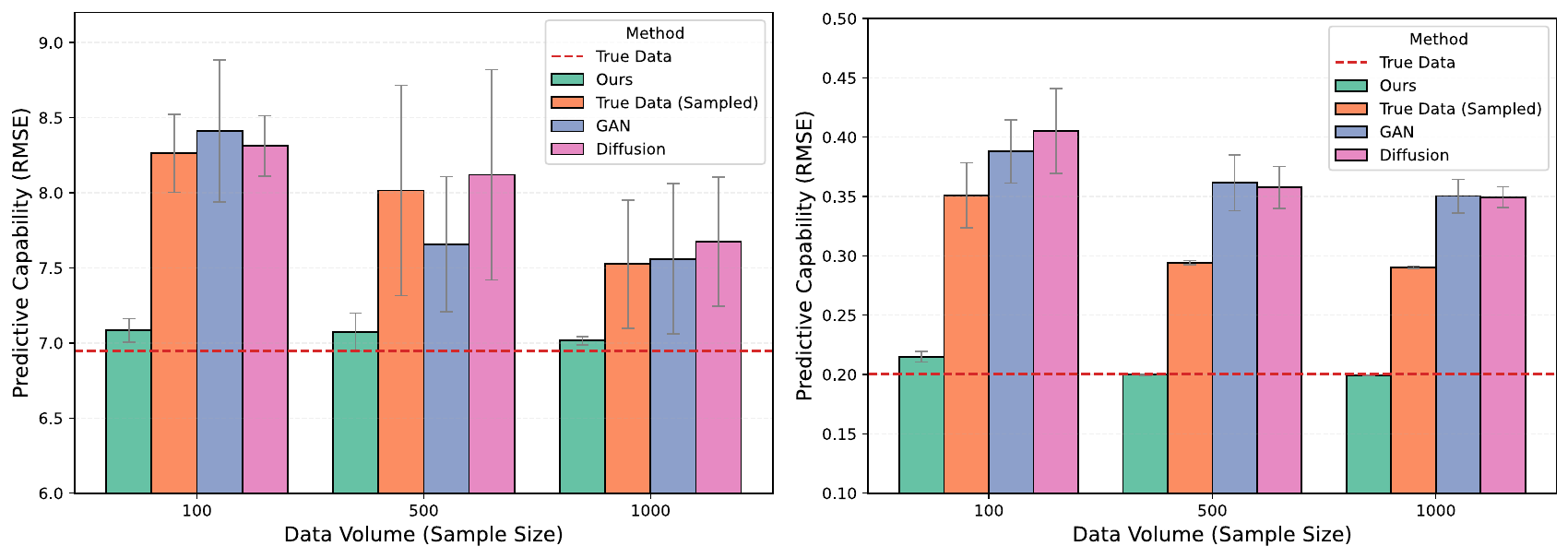}} 
	{Accuracy-Volume Results on Monte Carlo Datasets \label{figure MonteCarlo Accuracy-Volume}}
	{This figure reports the mean predictive capability (RMSE) of synthetic data generated by our framework and benchmark methods, evaluated across 10 independent runs under varying data volumes (sample sizes). Left: CLV Prediction Data; Right: Friedman Synthetic Regression Data. Error bars indicate the standard deviation of RMSE.}
\end{figure}

To evaluate the sample efficiency, we examine the RMSE of inference models trained on synthetic data generated by our framework, relative to that of two representative benchmark generative methods, GAN and Diffusion models, across three constrained data volumes: 100, 500, and 1000. For reference, the corresponding performance of original  data and original  data (Sampled) is also reported, where \textit{original  data} refers to the model trained on the full original dataset, while \textit{original  data (Sampled)} corresponds to the model trained on randomly sampled subsets matching the three specified sample sizes. Figure \ref{figure MonteCarlo Accuracy-Volume} shows the results on two Monte Carlo datasets. 

On the CLV Prediction dataset (left panel), the inference model trained on the full dataset (original  data) yields an RMSE of 6.948, which serves as a reference for the best possible performance. When the sample size reaches 1,000, original  data (Sampled) achieves an RMSE of 7.525. The GAN and Diffusion methods yield results very close to the original  data (Sampled), with RMSEs of 7.560 and 7.675, respectively. In contrast, our framework achieves a significantly lower RMSE of 7.016, highlighting its superior predictive performance under limited data. This advantage becomes even more pronounced as the data volume decreases. At a sample size of 100, all three benchmark methods suffer substantial performance degradation: GAN exhibits the largest increase in RMSE ($ + 0.852 $), followed by original  data ($ + 0.737 $) and Diffusion ($ + 0.636 $).  In comparison, our method maintains a remarkably stable performance, with only a minor increase of $+0.068$. Remarkably, using only 1\% of the data (sample size = 100), our method achieves predictive performance comparable to that of original  data (sample size = 10,000), demonstrating superior sample efficiency. These findings are consistent with the results observed on the Friedman Synthetic Regression dataset (right panel).
% where our method achieves an RMSE of 0.215 with only 100 samples—comparable to original  data (RMSE = 0.200 with 5,000 samples), and significantly lower than those of the three benchmark methods even at a sample size of 1,000 (original  data (Sampled): 0.290; GAN: 0.350; Diffusion: 0.349). 
These results highlight the remarkable sample efficiency of our framework, whereby strong predictive performance is attainable even with significantly reduced data volume. This efficiency offers notable practical benefits in real-world applications, including faster data transfer speed, lower storage requirements, and reduced computational overhead.

\subsubsection{Utility}
We evaluate the utility of our framework from three perspectives: stability, scalability, and velocity.

$\bullet$ \textbf{Stability: } 
In addition to predictive performance and data volume, Figure \ref{figure MonteCarlo Accuracy-Volume} also presents the stability results of our framework compared to benchmark approaches, measured by the standard deviation of RMSE over 10 repeated experiments. It can be observed that across all sample sizes and both datasets, our approach consistently achieves significantly lower RMSE standard deviations compared to the benchmark methods. For instance, on the CLV Prediction dataset (left panel) with a sample size of 1,000, the standard deviations for original  data (Sampled), GAN, and Diffusion methods are 0.427, 0.500, and 0.429, respectively, whereas our framework yields a remarkably low value of just 0.028. This exceptional stability enhances the reliability of downstream inference models trained on the synthetic data, making our approach more trustworthy in practical deployments.

$\bullet$ \textbf{Scalability: } 
To examine how the training time of our model scales with the number of training samples, we generate two synthetic datasets for each Monte Carlo setting, containing 1,000 and 100,000 observations, respectively, and train our model on each. 
We fix the batch size at 64 during training and observe that the per-iteration training time remains stable across different dataset sizes within each Monte Carlo setting.
Specifically, for the CLV Prediction dataset, the per-iteration times are 35.6 milliseconds and 37.0 milliseconds for datasets with 1,000 and 100,000 observations, respectively. For the Friedman Synthetic Regression dataset, the corresponding times are 19.8 milliseconds and 21.9 milliseconds. These results indicate that our framework scales efficiently with dataset size. This efficiency arises from the use of stochastic gradient descent (SGD), where model updates are performed on a subset of the data (i.e., the mini-batch), making the per-iteration computational cost dependent on the batch size rather than the overall dataset size.

\begin{figure}[h]
	\FIGURE
	{\includegraphics[width=0.8\linewidth]{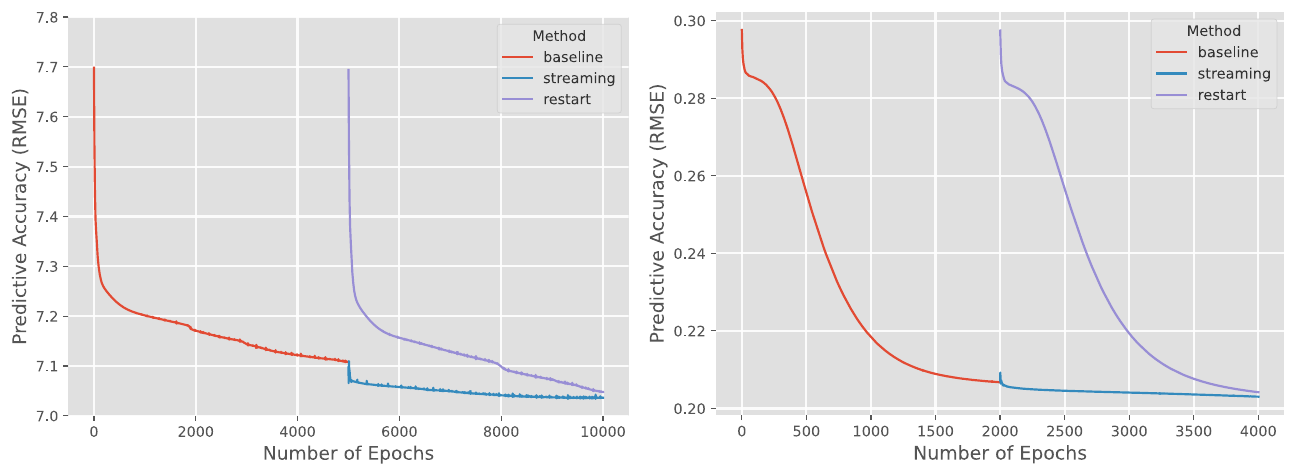}} 
	{Data Velocity Scalability Results on Monte Carlo Datasets \label{figure MonteCarlo Scalability}}
	{}
\end{figure}
$\bullet$ \textbf{Velocity: } 
We assess the data velocity of our framework by simulating a scenario with incrementally arriving data. Specifically, we compare two strategies for model updating: (1) ``restart'', where the model is retrained from scratch upon receiving new data; (2) ``streaming'', where the model is updated continuously using newly arrived data based on previously learned parameters (i.e., ``baseline'' model). Figure \ref{figure MonteCarlo Scalability} presents the RMSE trends across training epochs for both Monte Carlo datasets. We observe that the streaming approach achieves predictive performance comparable to that of the restart method, but with significantly fewer training epochs after new data arrives. These results highlight the advantage of leveraging the online nature of SGD, which allows our framework to incorporate newly arriving data in a computationally efficient manner without retraining from scratch.

\subsubsection{Extensibility in Terms of Data Augmentation}
\begin{figure}[h]
	\FIGURE
	{\includegraphics[width=0.8\linewidth]{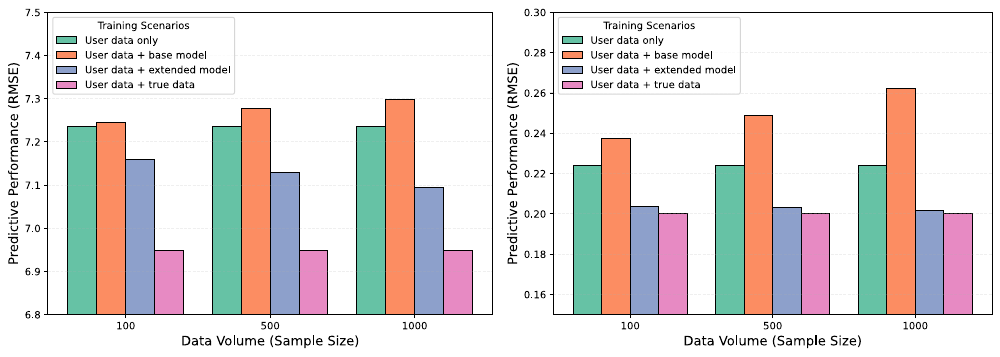}} 
	{Statistical Alignment Results on Monte Carlo Datasets \label{figure MonteCarlo Statistics-Alignment}}
	{This figure reports the predictive performance of inference models trained under four data scenarios: using only the user’s own data; user’s own data + synthetic data from the base model;
    user’s own data + synthetic data from the extended model (\ref{objective Wasserstein}); user’s own data + the original  data used to
    train the generator. Left: CLV Prediction Data; Right: Friedman Synthetic Regression Data.}
\end{figure}

To evaluate the data augmentation capability of the extended model (\ref{objective Wasserstein}), Figure \ref{figure MonteCarlo Statistics-Alignment} reports the predictive performance of inference models trained under four data scenarios across three different data volumes. Specifically, we simulate a user's own dataset by holding out 3,000 samples from the original dataset, while the remaining samples are used to train our generative models.
As expected, the scenario that augments the user data with the remaining real data consistently achieves the lowest RMSE, serving as an empirical best bound due to the availability of both large and high-quality training data. 
Compared to using the user's own data alone, augmenting with synthetic data from the base model leads to an increase in RMSE, indicating distributional misalignment and degraded predictive performance. In contrast, augmentation with synthetic data from the extended model results in a reduction in RMSE, demonstrating that the extended model effectively improves distributional fidelity and thus enhances the predictive accuracy based on the user’s private dataset.

Additionally, to provide an intuitive illustration of how the regularization term in our model (\ref{objective Wasserstein}) influences the statistical alignment of the synthetic data, we examine the evolution of three distributional metrics with respect to the hyperparameter $\lambda$. The corresponding results are visualized and presented in the supplementary material (Appendix F). Overall, the observed trend confirms that as $\lambda$ increases, the values of the metrics decline rapidly and eventually stabilize at low levels, indicating enhanced statistical alignment between the synthetic and original data. These findings validate the flexibility and controllability of our extended model (\ref{objective Wasserstein}) in achieving statistical alignment, thereby enabling it to accommodate diverse user needs and task-specific requirements.

\subsubsection{Extensibility in Terms of Model Adaptability}
To evaluate the inference model adaptability of the synthetic data produced by our extended model (\ref{objective final}), we compare the predictive performance of KRR and SVR with the designated model, and show the results in Table \ref{tab: Model Specificity} (Synthetic data (Extended model)). It can be observed that both KRR and SVR achieve RMSEs close to that of the designated model. For example, when trained on Synthetic data (Extended model), KRR and SVR yield RMSEs of 7.124 and 7.138, respectively, both 
comparable to the results of the designated model (RMSE = 7.083). These results demonstrate the strong model adaptability of our extended model, allowing non-designated models to achieve predictive performance comparable to that of the designated model. This property enhances the practical flexibility and reusability of the synthetic data across diverse inference models.

\subsection{Real-world Data  Experiments}
To comprehensively evaluate the effectiveness of our proposed framework, we conduct experiments on three real-world customer-level behavioral datasets. Each dataset is processed to support customer-level outcome prediction based on historical behavioral features. 
% In addition, two widely used benchmark datasets from the UCI repository are included in the experiments to demonstrate the applicability of our framework beyond marketing-related scenarios.
\subsubsection{Datasets}
We provide a brief description of the three real-world marketing datasets below. For more details, please refer to the supplementary materials (Appendix E).

$\bullet$ \textbf{Instacart Online Grocery Shopping Dataset:}
The Instacart dataset contains online grocery purchase records from over 200,000 users. We construct a short-term demand forecasting task, where the goal is to predict the number of products in a user’s next order based on their historical purchasing behavior. A subset of 10,000 users is used, with 5,000 for training and 5,000 for testing.

$\bullet$ \textbf{Retailrocket Recommender System Dataset:}
The Retailrocket dataset includes interaction events (views, cart additions, purchases) of $140,000$ users on an e-commerce platform. We formulate a CLV prediction task, aiming to estimate the total number of purchases per user based on aggregated behavioral features. A total of 6,000 users are used for training, with the remaining users for testing.

$\bullet$ \textbf{CDNOW Customer Transaction Dataset:}
The CDNOW dataset contains timestamped transaction data from 23,570 customers collected between 1997 and 1998. We define a CLV prediction task where the target is each customer’s spending in a 6-month holdout period, based on behavioral features extracted from a prior 12-month calibration window. We sample 4,000 customers for training and use the rest for testing.

% $\bullet$ \textbf{Online News Popularity Dataset:} 
% The Online News Popularity dataset contains article-level metadata for 39,644 news items published by the Mashable news platform. Each article is described by a set of hand-crafted features, including content attributes (e.g., word count, number of images, presence of keywords), metadata (e.g., publication day, channel category), and temporal features. The prediction target is defined as the number of shares an article receives on social media, reflecting its popularity.  We randomly sample 20,000 articles for training and retain the remainder for testing.

% $\bullet$ \textbf{Wine Quality Dataset:}
% The Wine Quality dataset consists of physicochemical and sensory data for 6,497 samples of red and white Vinho Verde wines from Portugal. Each wine sample is described by 11 physicochemical attributes such as acidity, residual sugar, pH, alcohol content, and sulfur dioxide levels. The prediction target is the wine’s quality score, rated on a scale from 0 to 10, based on blind tastings by professional wine tasters. We randomly sample 5,000 instances for training and retain the rest for testing.

\subsubsection{Rivalry} 
Below, we present the empirical results evaluating the rivalry of our framework across three real-world datasets.

\begin{figure}[h]
	\FIGURE
	{\includegraphics[width=0.99\linewidth]{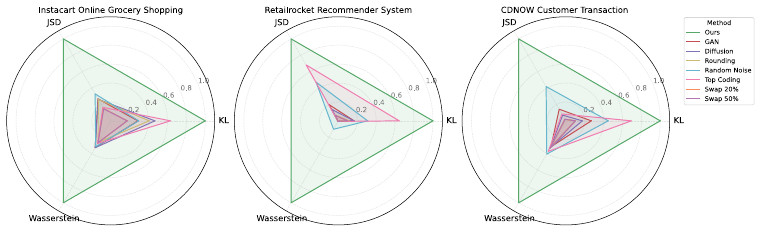}} 
	{Statistical Divergence Results on Real-world Datasets (Higher Is Better) \label{figure RealWorld Statistics-Divergence}}
	{This radar chart reports the KL divergence, JSD  and Wasserstein distance between the original dataset and the synthetic dataset generated by our base model and benchmark methods on real-world datasets.}
\end{figure}
$\bullet$ \textbf{Rivalry with Respect to External Data: } we evaluate the rivalry with respect to external data using statistical divergence and resistance to external augmentation, respectively.
Figure \ref{figure RealWorld Statistics-Divergence} presents the statistical divergence results of our base model and the baseline methods across real-world datasets, visualized using radar chart of the three distributional metrics. All metric values are normalized by their respective maxima to facilitate direct visual comparison. It can be observed that on all datasets, our method consistently achieves significantly higher values across all three metrics compared to the baseline methods, indicating that the synthetic data generated by our base model exhibits pronounced distributional differences from the original data.

We evaluate the resistance of our synthetic data to external augmentation across three real-world datasets by examining whether the introduction of additional user-owned samples into the synthetic dataset improves or degrades the performance of inference models. Following the setup used in the Monte Carlo experiments, we compare the predictive accuracy of inference models trained on the original synthetic data generated by our base model with its augmented version that includes 1,000 additional user-owned samples. The results are reported in Table \ref{tab: Real-world External Augmentation}. Across all three real-world datasets, we observe that introducing external user-owned samples into our synthetic data consistently leads to higher RMSEs, indicating a degradation in predictive performance. These results highlight the strong resistance of our synthetic data to external augmentation, effectively limiting unauthorized attempts to enhance model performance through  external data injection. This resistance empowers firms to maintain control over the use of shared data and reduce the risk of secondary development or unauthorized commercial exploitation.
\begin{table}[h]
	\TABLE
	{Resistance to External Augmentation Results on Real-world Datasets\label{tab: Real-world External Augmentation}}
	{	\begin{tabular}{ccccccc}
			%\hline
			\toprule
			\vspace{-2.5mm}
			& \multicolumn{2}{c}{Instacart} & \multicolumn{2}{c}{Retailrocket}& \multicolumn{2}{c}{CDNOW}\\ 
			
			\multirow{2}{*}{Sample size}& \hspace{0.05cm}\hrulefill\hspace{-0.6cm} & \hspace{-0.2cm}\hrulefill\hspace{0.05cm} &\hspace{0.05cm}\hrulefill\hspace{-0.6cm} & \hspace{-0.2cm}\hrulefill\hspace{0.05cm}
            &\hspace{0.05cm}\hrulefill\hspace{-0.6cm} & \hspace{-0.2cm}\hrulefill\hspace{0.05cm}\\
			& Synthetic data & Augmented data & Synthetic data & Augmented data	
            & Synthetic data & Augmented data
			\\\hline
			$ 100 $  & 6.033 & 6.217 & 1.781 &  3.632 &110.7  &  119.1
			\\
			$ 500 $  & 5.908 & 6.027 & 1.320  &  2.655 &  110.2&  116.4
			\\
			$ 1000 $  & 5.907 & 5.933  & 1.373 &  2.213 &  110.0&  114.0
			\\ \hline
			\toprule
	\end{tabular}}{This table reports the RMSE of our designated inference model under two settings: (1) trained on synthetic data generated by our base competitive model; and (2) trained on augmented data, which combines the synthetic data with additional user-owned samples.}
\end{table}

$\bullet$ \textbf{Rivalry with Respect to Inference Model: }
to assess inference model specificity on three real-world datasets, we compare the RMSEs of the designated inference model against those of KRR and SVR trained on synthetic data generated by our base rivalry model. The results, summarized in Table~\ref{tab: Real-world Model Specificity} (Synthetic data (Base model)), show that both KRR and SVR yield significantly higher RMSEs than the designated model. This indicates strong model specificity in the synthetic data, which impairs the generalization ability of non-designated models and thus helps prevent unauthorized model usage.

\begin{table}
	\TABLE
	{Inference Model Specificity and Adaptability Capability Results on Real-world Datasets\label{tab: Real-world Model Specificity}}
	{	\begin{tabular}{ccccccc}
			%\hline
			\toprule
			\vspace{-2.5mm}
			& & & \multicolumn{2}{c}{Synthetic data (Base model)} & \multicolumn{2}{c}{Synthetic data (Extended model)}\\ 
			
			\multirow{2}{*}{Dataset}&\multirow{2}{*}{Sample size}&\multirow{2}{*}{Designated model}& \hspace{0.05cm}\hrulefill\hspace{-0.6cm} & \hspace{-0.2cm}\hrulefill\hspace{0.05cm} &\hspace{0.05cm}\hrulefill\hspace{-0.6cm} & \hspace{-0.2cm}\hrulefill\hspace{0.05cm} \\
			&  &  & ~~~~~~~~KRR & SVR& ~~~~~~~~~~~KRR & SVR		
			\\\hline
			\multirow{3}{*}{Instacart}    & $ 100 $ &5.987  & ~~~~~~~~  6.892& 7.755 &  ~~~~~~~~~~~ 6.058&  6.186
			\\
			& $ 500 $ & 5.883 &  ~~~~~~~~ 7.060& 8.125 &  ~~~~~~~~~~~5.988&  5.965
			\\
			& $ 1000 $ & 5.883 & ~~~~~~~~ 6.807 &8.200  & ~~~~~~~~~~~ 5.970 &  5.935
			\\\cline{1-7}
			\multirow{3}{*}{Retailrocket}  & $ 100 $ & 2.112 &  ~~~~~~~~ 7.221&6.581 & ~~~~~~~~~~~  2.561 &  2.805
			\\
			& $ 500 $ & 1.966 &  ~~~~~~~~ 6.687&5.587  & ~~~~~~~~~~~1.822 &  1.990
			\\
			& $ 1000 $ & 2.086 & ~~~~~~~~5.862  & 5.042  & ~~~~~~~~~~~1.758 & 2.054   
			\\\cline{1-7}
			\multirow{3}{*}{CDNOW}  & $ 100 $ & 111.7 &  ~~~~~~~~ 129.5& 126.8&  ~~~~~~~~~~~ 112.1 &  121.5
			\\
			& $ 500 $ &112.8  &  ~~~~~~~~ 123.2& 120.3 & ~~~~~~~~~~~ 110.6&  113.3
			\\
			& $ 1000 $ & 112.9 & ~~~~~~~~  121.6& 118.6  & ~~~~~~~~~~~ 109.5&111.4
			% \\\cline{1-7}
			% \multirow{3}{*}{Online News}  & $ 100 $ & 574.2 &  ~~~~~~~~ 581.3& 603.6&  ~~~~~~~~~~~573.5  &  580.8
			% \\
			% & $ 500 $ & 573.9 &  ~~~~~~~~ 594.0&  600.3& ~~~~~~~~~~~574.1 &  578.0
			% \\
			% & $ 1000 $ & 573.8 & ~~~~~~~~  590.7& 593.4  & ~~~~~~~~~~~573.8 &576.6
			% \\\cline{1-7}
			% \multirow{3}{*}{Wine Quality}  & $ 100 $ & 0.703 &  ~~~~~~~~ 0.762&0.850 &  ~~~~~~~~~~~  0.731&  0.729
			% \\
			% & $ 500 $ & 0.698 &  ~~~~~~~~ 0.768& 0.821 & ~~~~~~~~~~~ 0.727&  0.722
			% \\
			% & $ 1000 $ &0.697  & ~~~~~~~~ 0.745 &  0.823 & ~~~~~~~~~~~ 0.727&0.721
			\\ \hline
			\toprule
	\end{tabular}}{This table reports the RMSE of our designated inference model, compared against two alternative scenarios: (1) KRR and SVR trained on synthetic data generated by our base competitive model; and (2) KRR and SVR trained on synthetic data generated by our extended model for inference model adaptability (see extended model  (\ref{objective final}).}
\end{table}

\begin{figure}[h]
	\FIGURE
	{\includegraphics[width=0.99\linewidth]{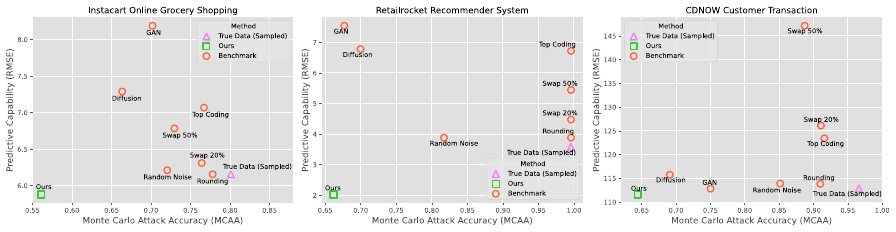}} 
	{Accuracy-Privacy Results on Real-world Datasets \label{figure RealWorld Accuracy-Privacy}}
	{This figure reports the predictive capability (RMSE) and Monte Carlo Attack Accuracy (MCAA) for our framework and benchmarks on the three real-world datasets, with a sample size of 3,000. A lower MCAA indicates stronger privacy protection.}
\end{figure}

$\bullet$ \textbf{Privacy Protection: }
We evaluate the balance between predictive accuracy (measured by RMSE) and privacy protection (measured by MCAA) for our framework and baseline methods on real-world datasets, and present the results in Figure \ref{figure RealWorld Accuracy-Privacy}. The results reveal a clear tradeoff among the baseline methods: perturbation-based techniques (e.g., Rounding, Topcoding, Random Noise, and Swapping) tend to maintain relatively high predictive accuracy (low RMSE) but offer weaker privacy protection (high MCAA), while generative models like GAN and Diffusion provide improved privacy (lower MCAA) at the cost of degraded prediction performance (higher RMSE). In contrast, our framework consistently occupies the bottom left region of the RMSE-MCAA plots across all datasets, indicating a superior balance between accuracy and privacy.

For example, on the Instacart Online Grocery Shopping dataset, we observe that among the baseline methods, the original  data (Sampled) achieves the highest predictive accuracy with an RMSE of 6.15, but this comes at the cost of privacy, reflected in the highest MCAA of 0.801 due to the lack of any privacy-preserving approach.  On the other hand, the best privacy-preserving baseline is the Diffusion model, which achieves the lowest MCAA of 0.663, but suffers from a significant drop in predictive accuracy, with an RMSE of 7.29. In contrast, our framework outperforms all baselines on both fronts: it achieves a lower RMSE of 5.88, indicating superior predictive accuracy, and a substantially reduced MCAA of 0.561, which represents over a 16\% improvement in privacy protection compared to Diffusion. These findings highlight that our synthetic data maintains high predictive accuracy while significantly reducing privacy leakage risk. This allows firms to safely share customer behavioral data without compromising individual privacy, while enabling data users to reliably forecast customer demand and make informed marketing decisions.

\subsubsection{Data Volume} 

\begin{figure}[h]
	\FIGURE
	{\includegraphics[width=0.99\linewidth]{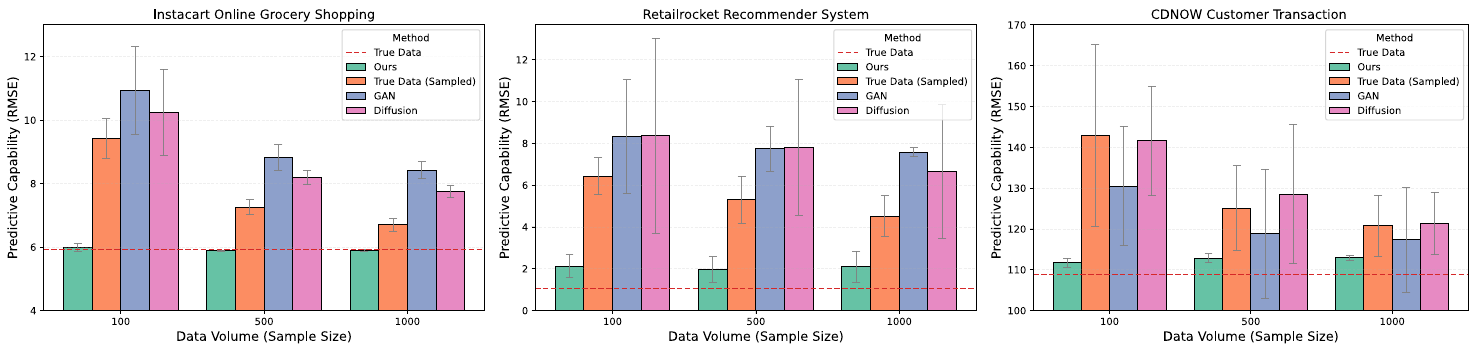}} 
	{Accuracy-Volume Results on Real-world Datasets \label{figure RealWorld Accuracy-Volume}}
	{This figure reports the mean predictive capability (RMSE) of synthetic datasets generated by our framework and benchmarks, evaluated across 10 independent runs under varying data volumes (sample sizes). Error bars indicate standard deviation of RMSE.}
\end{figure}

We compare the predictive performance of our proposed framework against three baselines—original  data (Sampled), GAN, and Diffusion—under varying data volumes. Figure~\ref{figure RealWorld Accuracy-Volume} presents RMSE results for sample sizes of 100, 500, and 1,000, with original  data results included as references for upper-bound performance. Across all datasets and sample sizes, our framework consistently demonstrates superior predictive accuracy to the three baselines, with its advantage becoming increasingly pronounced as the data volume decreases. Furthermore, our framework exhibits consistently lower standard deviations in RMSE across runs,  indicating greater stability and enhanced reliability for real-world scenarios.

For instance, on the Instacart Online Grocery Shopping dataset, where the prediction task involves estimating the number of products in a customer’s next order, our framework achieves substantial improvements across all data volume settings. At the smallest sample size of 100, our method achieves an RMSE of 5.99, which is comparable to that of original  data (5.926 with 10,000 samples), and substantially outperforms original  data (Sampled) (9.42), GAN (10.94), and Diffusion (10.24). This represents a reduction of over 35\% in prediction error relative to the best-performing baseline. Even as the sample size increases to 1000, our method maintains a clear advantage, achieving an RMSE of 5.88  compared to 6.70 for the best baseline (original  data (Sampled)),  representing a further reduction of more than 12\%. These results suggest that inference models trained on our synthetic data can deliver highly accurate predictions of short-term customer demand at a relatively low data cost. From a marketing perspective, such predictive fidelity enables sellers to better anticipate short-term demand, optimize inventory levels, and personalize product recommendations. Ultimately, this supports more informed operational planning and customer engagement strategies, even in scenarios with limited access to real user data. 

\subsubsection{Extensibility in Terms of Data Augmentation} 

\begin{figure}[h]
	\FIGURE
	{\includegraphics[width=0.99\linewidth]{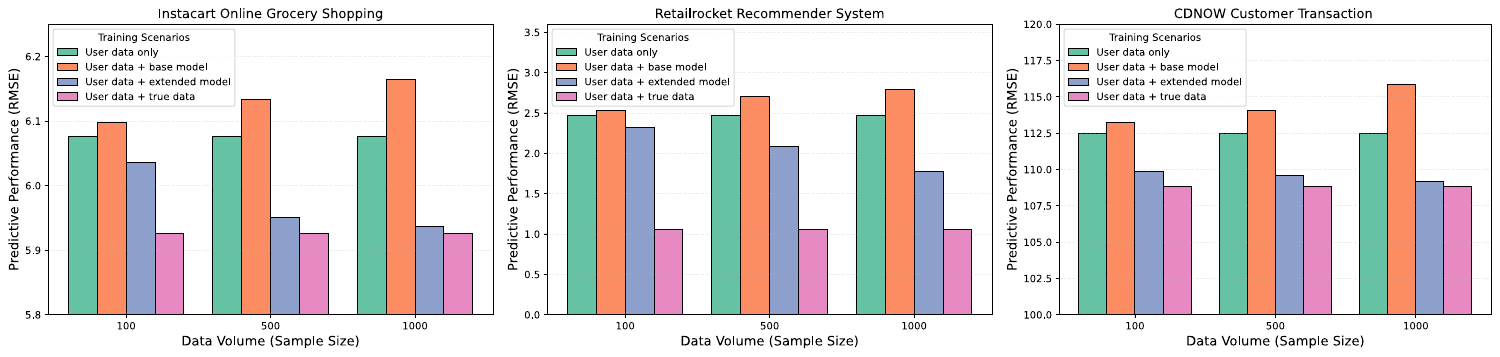}} 
	{Statistical Alignment Results on Real-world Datasets \label{figure RealWorld Statistics-Alignment}}
	{This figure reports the predictive performance of inference models trained under four data scenarios: using only the user’s own data; user’s own data + synthetic data from the base model;
    user’s own data + synthetic data from the extended model (\ref{objective Wasserstein}); user’s own data + the original  data used to
    train the generator.}
\end{figure}

Figure \ref{figure RealWorld Statistics-Alignment} reports the predictive performance of inference models trained under four data scenarios, providing an empirical evaluation of the data augmentation
capability of the extended model (\ref{objective Wasserstein}). We observe that, compared to training the inference model using only the user data, augmenting with synthetic data from the base model leads to an increase in RMSE, while augmenting with data from the extended model results in a reduction in RMSE. This contrast highlights the extended model’s enhanced statistical alignment relative to the base model, which in turn enables more effective data augmentation and enhanced predictive performance. We also visualize the behavior of the three statistical metrics in the supplementary material (Appendix F) and observe a rapid decline in their values as $\lambda$ increases,  further validating the statistical alignment capability of the extended model. This flexible statistical alignment and data augmentation capability empower firms with fine-grained control over the informativeness of synthetic data, allowing differentiated data customization for varying levels of user demand and marketing value.

\subsubsection{Extensibility in Terms of Model Adaptability} 
We evaluate inference model adaptability by comparing the RMSE of the designated model with those of KRR and SVR trained on synthetic data generated by the extended model (\ref{objective final}), and report the results in Table~\ref{tab: Real-world Model Specificity} (Synthetic data (Extended model)). Contrary to the results of the base model, when trained on synthetic data from the extended model, both KRR and SVR achieve much closer RMSE values to the designated model. This demonstrates that the extended model effectively improves model adaptability, enabling a broader range of inference models to achieve similar predictive performance.
These findings highlight the strategic flexibility of our framework in data-sharing contexts: firms can suppress predictive performance for unauthorized inference models using the base model to prevent misuse, or promote broader compatibility through the extended model for trusted collaborators.

\section{Conclusion}\label{sec:Conclusion}
This paper addresses a fundamental tension in data sharing: while collaboration can create significant value, firms often hesitate to share proprietary data due to competitive risks. We introduced a novel data synthesis framework that transforms this strategic dilemma by enabling controlled utility distribution.
Our two-stage approach—Knowledge Encapsulation and Asymmetric Utility Forging—allows data providers to precisely control who can extract value from shared data. By optimizing synthetic datasets for specific algorithmic ``keys", we create data artifacts that reveal their full utility only to designated recipients while remaining obfuscated to unauthorized users. This approach fundamentally shifts the paradigm from statistical mimicry to strategic utility design.

Extensive experiments across Monte Carlo simulations and real-world datasets demonstrate that our framework achieves superior sample efficiency, requiring only 5\% of the original data size while maintaining comparable predictive performance. The synthetic data exhibits strong competitiveness through statistical divergence, model specificity, and resistance to external augmentation, while flexible extensions enable controllable relaxation of these constraints based on specific sharing requirements.
The practical implications are significant. Our framework provides firms with a technical solution to engage in value-creating data collaboration while maintaining competitive control. This capability is particularly relevant as regulatory frameworks like the EU Data Act mandate greater data sharing, requiring mechanisms that balance innovation with corporate autonomy.

Several promising directions emerge for future research. First, extending the framework to support multi-party data sharing scenarios where multiple firms contribute data while maintaining individual competitive advantages. Second, developing adaptive mechanisms that automatically adjust utility distribution based on evolving trust relationships and market conditions. Third, investigating the application of our approach to unstructured data types such as text and images, which present unique challenges for controllable utility design. Finally, exploring the integration of federated learning principles to enable collaborative model training without direct data sharing, further reducing privacy and competitive risks.
These extensions could establish a comprehensive ecosystem for strategic data collaboration, transforming how firms approach the co-opetition dilemma in the digital economy.

%\THEEndNotes
\begingroup \parindent 0pt \parskip 0.0ex \def\enotesize{\normalsize} \theendnotes \endgroup

% Appendix here
% Options are (1) APPENDIX (with or without general title) or
%             (2) APPENDICES (if it has more than one unrelated sections)
% Outcomment the appropriate case if necessary
%
% \begin{APPENDIX}{<Title of the Appendix>}
% \end{APPENDIX}
%
%   or
%
% \begin{APPENDICES}
% \section{<Title of Section A>}
% \section{<Title of Section B>}
% etc
% \end{APPENDICES}

% Acknowledgments here
% \ACKNOWLEDGMENT{We would like to express our sincere gratitude to [acknowledge individuals, organizations, or institutions] for their invaluable contributions to this research. We are also grateful to [mention any additional acknowledgements, such as technical assistance, data providers, or colleagues] for their support and assistance throughout the course of this work.}

% References here (outcomment the appropriate case)

\bibliographystyle{informs2014}
\bibliography{ref1}

%%%%%%%%%%%%%%%%%
\end{document}